\documentclass[11pt]{article}

\usepackage[preprint]{acl}

\usepackage{times}
\usepackage{latexsym}

\usepackage[utf8]{inputenc}
\usepackage{textcomp}
\usepackage[T1]{fontenc}
\usepackage{natbib}
\usepackage{amsmath}
\usepackage{amssymb}
\usepackage{booktabs}
\usepackage[table]{xcolor}
\usepackage{multicol}
\usepackage{multirow}
\usepackage{algorithm}
\usepackage{algpseudocode}
\usepackage[table]{xcolor}

\usepackage[utf8]{inputenc}

\usepackage{microtype}

\usepackage{inconsolata}
\usepackage{subcaption}
\usepackage{booktabs,tabularx,array}

\usepackage{graphicx}
\usepackage{xspace}
\usepackage{fvextra}
\usepackage{amsmath}

\title{Self-Improvement with Auxiliary Rewards: \\ Learning from Verified Traces and Evaluating Transfer}

\author{Minju Gwak \quad Alan Ritter \\
  Georgia Institute of Technology \\
  \texttt{mgwak9@gatech.edu, alan.ritter@cc.gatech.edu} \\}

\begin{document}
\maketitle
\begin{abstract}
Answer-level rewards cannot distinguish reasoning traces with equally correct outcomes. We study whether auxiliary trajectory preferences improve small-language-model self-improvement and transfer under limited supervision. We compare confidence-based (SC) and entropy-based metrics (UID-Conj.), which combines global entropy variation and local smoothness, through online reward augmentation and offline training on correct responses. Online experiments adapt Qwen3-1.7B on 30 AIME problems without reference answers; offline experiments use historical AIME problems. UID-Conj. improves the online macro-average across the adaptation benchmark and three transfer benchmarks by 2.08 percentage points over unmodified TTRL. The strongest evidence comes from offline weighted SFT: UID-Conj. exceeds uniform weighting on the same response pool by 1.93 points, with a paired prompt-bootstrap 95\% CI of [0.44, 3.44].  Benefits are task-dependent: SC rewards judged trace quality and instruction following, while UID-Conj. rewards yield stronger training outcomes in several settings. We find that auxiliary preferences can expose learning value among equally correct traces not captured during inference time, motivating their evaluation through downstream gains.
\end{abstract}

\section{Introduction}

Recent advances in reasoning post-training have strengthened small language models~\citep{yang2025qwen3,guo2025deepseek}, but further improvement under limited supervision requires making better use of available training problems. Self-improvement offers one route: models generate reasoning traces and learn from feedback on their own responses~\citep{zelikman2022star,haluptzok2023language,yuan2024selfrewarding}. Answer-level feedback, such as the checkable outcomes used in reinforcement learning with verifiable rewards (RLVR)~\citep{guo2025deepseek,wang2025reinforcement}, provides a practical training signal but cannot distinguish responses with equally correct final answers. Such responses may differ in the soundness of their reasoning, the completeness of their justification, and the amount of unnecessary repetition; they may also differ in their value as training examples. Auxiliary preferences over these traces could provide additional supervision without requiring new problems or step-level annotations. We therefore ask: \emph{when answer-level rewards cannot distinguish successful solutions, can auxiliary trajectory preferences improve learning and downstream transfer?}

\begin{figure}
    \centering
    \includegraphics[width=\linewidth]{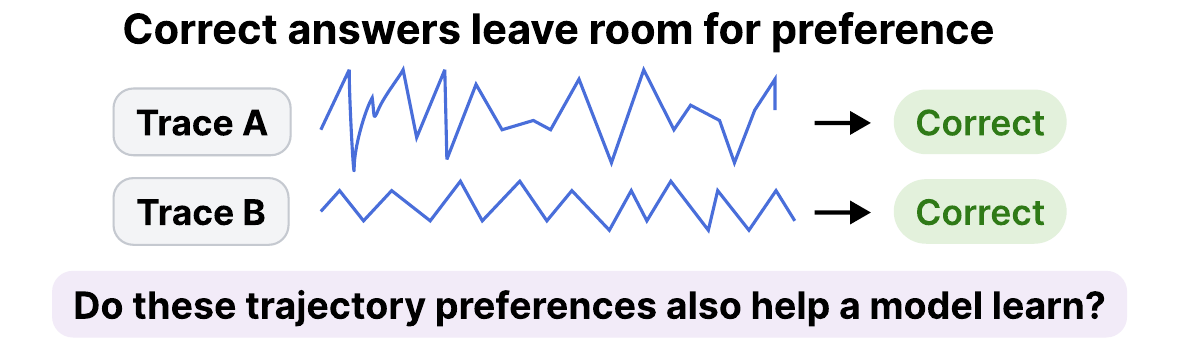}
    \caption{Correct answers leave room for preference. We investigate whether these trajectory preferences improve learning and downstream transfer.}
    \label{fig:motivating-fig}
    \vspace{-1.5em}
\end{figure}

We investigate two auxiliary signals derived from the model's own predictive token distributions. Self-Certainty (SC)~\citep{kang2025scalable} favors trajectories with greater predictive confidence. The second signal, inspired by uniform information density (UID)~\citep{meister-etal-2021-revisiting}, favors trajectories that combine global entropy variation with local smoothness, following observations of locally smooth but globally non-uniform information flow in successful reasoning~\citep{gwak-etal-2026-revisiting}. These signals define alternative preferences among responses receiving the same answer-level reward.

The central hypothesis of our study is that auxiliary preferences provide useful learning signals beyond answer-level feedback, improving task performance under limited supervision. To test this hypothesis, we compare training with and without auxiliary preferences while retaining the underlying answer-level supervision. Specifically, we adopt test-time reinforcement learning (TTRL)~\citep{zuo2026ttrl} for only 2 epoch of training to adapt the model on a small set test time without reference answers, therefore using only 30 AIME2026 problems. Within TTRL, we augment majority-agreement rewards with auxiliary preferences among responses supporting the majority answer, preserving the reward between agreeing and disagreeing responses.

Results show that auxiliary trajectory preferences can improve self-improvement from answer-level feedback. After adaptation on just 30 AIME2026 problems without reference answers, TTRL augmented with a Self Certainty and UID Conjunction metric (UID-Conj.) improve Qwen3-1.7B's macro-average accuracy across four mathematics and science benchmarks by 2.03 and 2.08 percentage points, respectively. The benefits extend to critique-verdict prediction and evidence-conditioned scientific reasoning. UID-Conj. improves CriticBench verdict accuracy by 1.24 points, incorrect-class F1 by 1.58 points, and LabBench2 accuracy by 0.76 points, without task-specific training. Complementary offline extensions show that UID-Conj. also improves Qwen3-1.7B's macro-average by 1.25 points over random correct--correct DPO preferences and by 1.93 points over uniform SFT.
To summarize, our contributions are:
\begin{itemize}
    \vspace{-0.1em}
    \item[$\circ$] An independent assessment of whether auxiliary signals distinguish reasoning quality among correct responses.
    \item[$\circ$] A controlled study of auxiliary trajectory preferences for small-model self-improvement with limited training data, spanning online and offline settings with controls matched within each objective.
    \item[$\circ$] An evaluation of transfer from mathematical self-improvement to critique and research-level reasoning.
    \vspace{-0.7em}
\end{itemize}

\begin{figure*}[t]
    \centering
    \includegraphics[width=\linewidth]{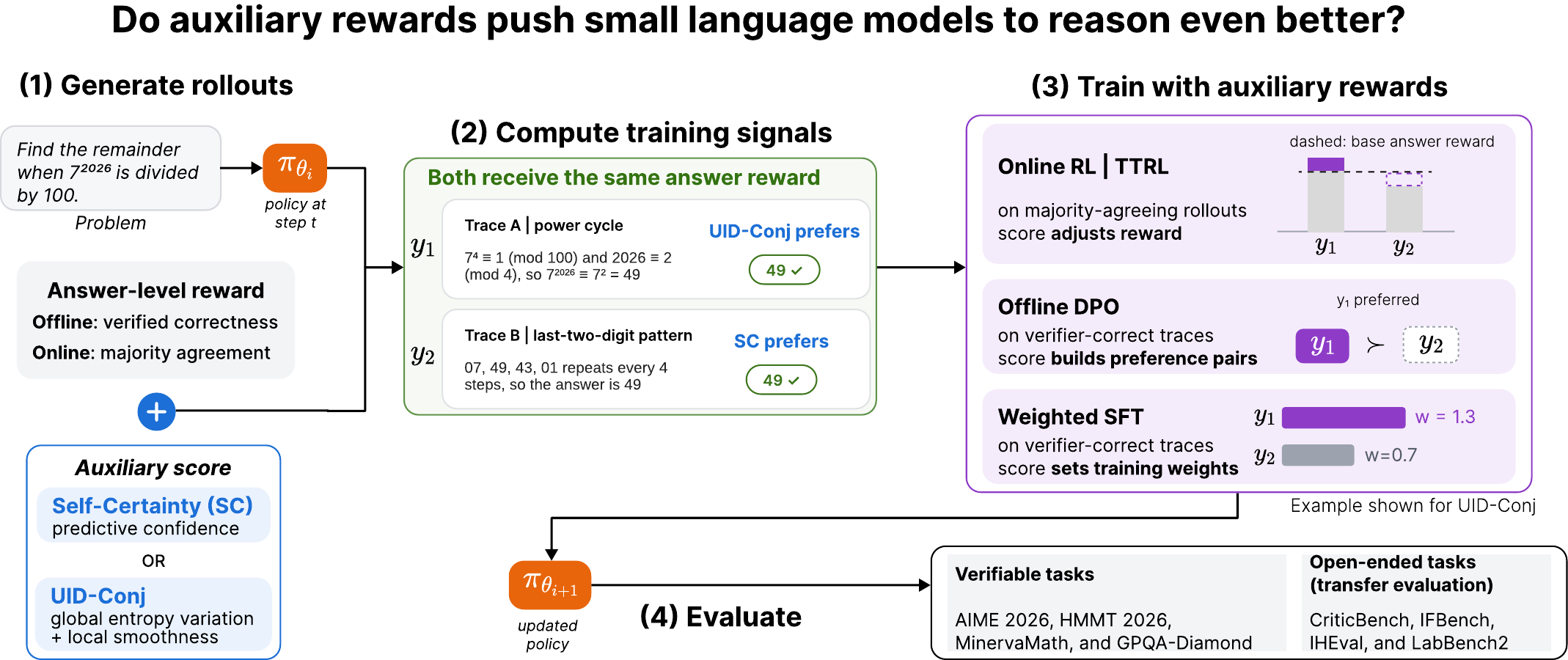}
    \caption{Self-improvement with auxiliary rewards. SC or UID-Conj distinguishes reasoning traces with equal answer-level rewards, guiding online TTRL reward augmentation, offline DPO preferences, or weighted SFT. The updated policy is evaluated on verifiable reasoning and open-ended transfer tasks.}
    \label{fig:placeholder}
    \vspace{-1.0em}
\end{figure*}
\section{Related Work}

\paragraph{Best-of-$N$ selection.}
Selection methods exploit different properties of sampled reasoning. Self-consistency aggregates agreement across final answers~\citep{wang2023selfconsistency}, whereas Self-Certainty (SC) scores responses using statistics of predictive distributions~\citep{kang2025scalable}. The uniform information density hypothesis concerns how information is distributed across language~\citep{meister-etal-2021-revisiting}; its application to LLM reasoning associates successful trajectories with local smoothness and global non-uniformity~\citep{gwak-etal-2026-revisiting}. These findings leave two questions unresolved: whether the scores identify useful differences among already-correct responses, and whether reinforcing those improves generation.

\paragraph{Self-improvement with outcome, process, and internal feedback.}
Self-improvement methods use feedback on model-generated responses to refine reasoning. STaR learns from its own rationales that yield correct answers~\citep{zelikman2022star}, while DeepSeek models use RL with outcome feedback~\citep{shao2024deepseekmath,guo2025deepseek}. Process supervision provides finer-grained feedback on intermediate steps~\citep{lightman2023lets}, while there have been research on preferring answers by length~\cite{arora2025training}. Closest to ours, Intuitor uses SC as the sole reward in GRPO and reports generalization from mathematics to out-of-domain tasks~\citep{zhao2026learning}, and ~\citet{agarwal2025unreasonable} minimizes entropy during training. This establishes internal confidence as a self-improvement signal and raises a specific question: does entropy-trajectory structure add useful supervision beyond confidence when correctness feedback is available?

\paragraph{Transfer beyond verifiable training tasks.}
Math datasets with known answers provide accessible supervision for reasoning models~\citep{hendrycks2021measuring,shao2024deepseekmath}, but the intended applications may exceed this supervision in both difficulty and scope. Easy-to-hard generalization studies whether supervision from simpler problems can support harder math reasoning~\citep{sun2024easytohard}, while scientific question answering exposes the difficulty of evaluating outputs that require specialized expertise~\citep{rein2023gpqa}. Open research problems often lack gold answers for outcome-based rewards, even when proposed solutions can be checked. This gap motivates testing both transfer to harder math and transfer beyond verified answer generation.
\section{Characterizing Trajectory Preferences: Do Auxiliary Signals Add Value beyond the Final Answer?}
\label{sec:preliminary-rollouts}

Before evaluating training utility, we characterize what the auxiliary signals select. We examine whether they improve answer selection and which reasoning properties they favor among verifier-correct responses. These diagnostics describe the preferences induced by each signal. We compare trajectory-based selection with and without answer aggregation, and examine changes relative to majority voting, and assess selected traces with a blinded external judge. Specifically, we ask:
\begin{enumerate}
    \vspace{-0.5em}
    \item Do auxiliary rewards complement answer-level agreement? (Section~\ref{sec:2.2})
    \item Do auxiliary rewards distinguish reasoning quality at fixed final-answer correctness? (Section~\ref{sec:2.4})
\end{enumerate}

\subsection{Experiment Setup}
\label{subsec:rollout-setup}

\paragraph{Models and rollout configurations.}
We evaluate fixed Qwen3-1.7B and Qwen3-4B checkpoints with thinking enabled, temperature 0.7, top-$p$ 0.95, and response and context limits of 8,192 and 16,384 tokens, respectively. We sample eight responses per prompt and evaluate all selectors on identical nested pools of size $n\in\{2,4,8\}$.

\paragraph{Datasets and prompts.}
We use AIME2026,
HMMT2026,
and MinervaMath,
for both inference-time analyses. See Appendix~\ref{app:prompt} for prompts.

\paragraph{Selection methods.}
We compare three selectors that directly selects among the rollout batch: \textbf{Majority Voting}, which selects the most frequent final answer. \textbf{Self-Certainty (SC)}, which selects the response with the highest confidence-based score, \textbf{UID-Conj}, which selects the response with the highest product of two calibrated global and local scores, favoring responses that score highly on both. See Appendix~\ref{app:selector-metrics} for how each selector is calculated. We combine each selector mentioned above with Borda Voting (Borda)~\citep{dwork2001rank}, an answer-level aggregation method that support final answers using response-rank weights.

\paragraph{Evaluation.}
We report mean accuracy over all prompts for each of the selectors, counting missing selections as incorrect.

\begin{table*}[t]
    \centering
    \captionsetup{width=\textwidth}
    \resizebox{0.9\textwidth}{!}{%
        \begin{tabular}{lc|cc|c c|cc|c}
            \toprule
            & \multicolumn{4}{c}{Direct selectors}
            & \multicolumn{4}{c}{Borda-added selectors} \\
            \cmidrule(lr){2-5}
            \cmidrule(lr){6-9}
            Model
            & \shortstack{Maj.}
            & SC
            & UID-Conj.
            & \shortstack{$\Delta$ best-worst\\(pp)}
            & \shortstack{Maj.}
            & SC + Borda
            & UID-Conj. + Borda
            & \shortstack{$\Delta$ best-worst\\(pp)} \\
            \midrule
            Qwen3-1.7B
            & \textbf{25.9 }& 25.4 & 25.7 & 0.78
            & 25.9 & 26.3 & \textbf{27.0} & 1.11 \\
            Qwen3-4B
            & 33.9 & 33.8 & \textbf{34.2} & 0.44
            & 33.9 & \textbf{34.7} & 34.2 & 0.78 \\
            \bottomrule
        \end{tabular}%
    }
    \caption{Accuracy (\%) averaged over rollout samples $n\in\{2,4,8\}$, over 300 pooled questions per model. $\Delta$ best-worst is the within-setup selector range in percentage points, including Majority Voting (Maj.)}
    \label{tab:prelim1-table}
    \vspace{-0.5em}
\end{table*}





\begin{figure}
    \centering
    \includegraphics[width=0.95\linewidth]{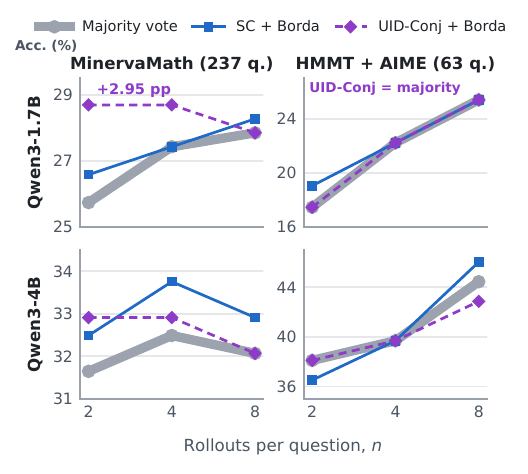}
    \caption{Mean accuracy of Auxiliary reward + Borda Voting vs. Majority Voting for Qwen3-1.7B (Top) and Qwen3-4B (Bottom), across rollout $n\in\{2,4,8\}$.}
    \label{fig:figure3_selectors}
    \vspace{-1.0em}
\end{figure}

\subsection{Auxiliary Rewards with Answer-level Aggregation}
\label{sec:2.2}
\label{subsec:preliminary-selection}

Direct selection using auxiliary scores does not reveal how the selected reasoning traces differ (Table~\ref{tab:prelim1-table}). We therefore compare trace quality while holding final-answer correctness fixed. Averaged across rollout budgets, SC and UID-Conj achieve 25.4\% and 25.7\% accuracy for Qwen3-1.7B, compared with 25.9\% for majority voting; for Qwen3-4B, the corresponding accuracies are 33.8\% and 34.2\%, versus 33.9\%. However, incorporating answer-level agreement through Borda aggregation yields improvements, with UID-Conj + Borda achieving the highest accuracy for Qwen3-1.7B at 27.0\% and SC + Borda performing best for Qwen3-4B at 34.7\%. These gains vary across rollout budgets and model size (Figure~\ref{fig:figure3_selectors}). For Qwen3-1.7B, auxiliary scores with Borda aggregation improve selection most clearly at smaller budgets. For Qwen3-4B, gains remain modest on MinervaMath, while results on HMMT and AIME depend on both the scoring method and rollout budget. Thus, auxiliary rewards provide complementary signals to answer-level agreement, but their selection benefits are context-dependent rather than uniform.

\subsection{Independent Assessment of Trace Quality}
\label{sec:2.4}
\label{subsec:preliminary-trace-quality}

The modest accuracy gains in Section~\ref{sec:2.2} do not substantially establish the effectiveness of the auxiliary scores. Also, it lacks a qualitative analysis on when which signal is more useful than the other. We therefore compare trace quality while holding final-answer correctness fixed. From each eligible candidate pool, we select verifier-correct responses using SC, UID-Conj, and random correct selection, then have a judge independently assess their validity, sufficiency, consistency, and relevance/non-redundancy without access to selector identity or a reference answer. Appendix~\ref{app:trace-quality} details the selection protocol, rubric, and aggregation procedure.

\begin{figure}
        \vspace{-0.5em}
        \includegraphics[width=\columnwidth]{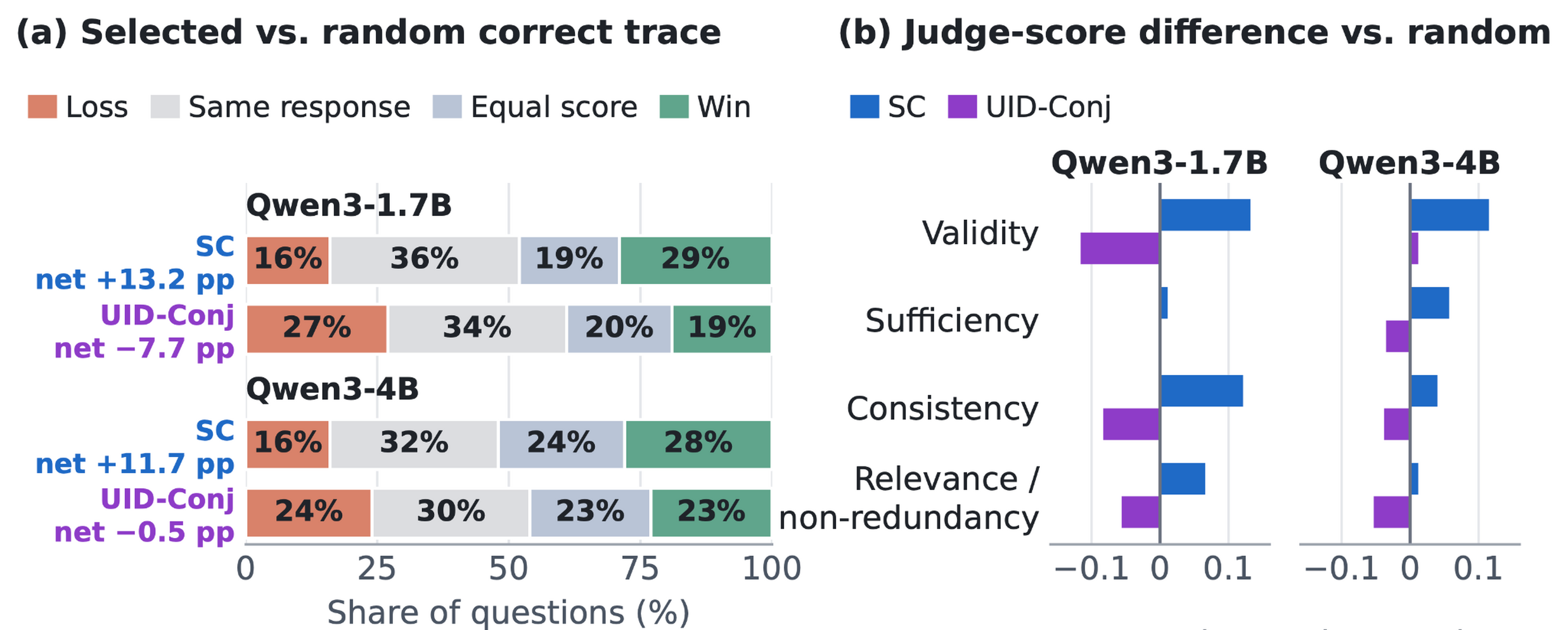}

        \caption{Trace quality for same answers. (Left) SC, UID-Conj. vs random correct selection, showing wins, losses, identical-response ties, and equal-score ties. Net denotes wins -- losses in pp. (Right) Mean differences on four reasoning criteria. Results average eligible rollout budgets within each question, then questions equally.}
        \label{fig:aux_vs_rand}
        \vspace{-1.0 em}
\end{figure}

\paragraph{Distinguishing good reasoning beyond answer correctness.}
SC selects higher-scoring traces than random correct selection, with net win rates of $13.2$ and $11.7$ percentage points for Qwen3-1.7B and Qwen3-4B, respectively (Figure~\ref{fig:aux_vs_rand}, left). UID-Conj instead yields $-7.7$ and $-0.5$ points. Although ties account for more than half of the outcomes, these results suggest that auxiliary scores can distinguish trace quality among correct responses, with benefits depending on the scoring.

\paragraph{Relative strengths of each signal.}
The auxiliary signals favor different aspects of trace quality: SC performs better on substantive reasoning criteria, including validity and consistency, whereas UID-Conj scores higher on relevance/non-redundancy despite weaker performance on most other criteria (Figure~\ref{fig:aux_vs_rand}, right). Overall, SC retains a mean total-score advantage of $0.584$ and $0.338$ points for Qwen3-1.7B and Qwen3-4B, respectively (Table~\ref{tab:judge-uid-vs-sc}). This contrast suggests that the signals may offer task-dependent benefits, motivating tests of whether SC is more useful across general tasks and UID-Conj's preference for less redundant traces benefits reasoning-intensive tasks. We next examine whether these differences in trace selection translate into differences in training effectiveness.

\begin{table}[t]
    \centering
    \small
    \captionsetup{width=\columnwidth}
    \resizebox{\columnwidth}{!}{%
        \begin{tabular}{lrrrrrr}
            \toprule
            Model
            & Quest.
            & Pairs
            & \shortstack{Net $\Delta$\\(pp)}
            & \shortstack{Total\\gap}
            & \shortstack{Large\\gap}
            & \shortstack{Same\\resp. (\%)} \\
            \midrule
            Qwen3-1.7B &  91 & 226 & -21.98 & -0.58 & -0.46 & 28.21 \\
            Qwen3-4B  & 107 & 295 & -15.42 & -0.34 & -0.27 & 30.69 \\
            \bottomrule
        \end{tabular}%
    }
    \caption{
        UID-Conj. vs SC on correct-only cohorts.
        Net $\Delta$ computed as UID-Conj -- SC; negative values
        favor SC (See Figure~\ref{fig:aux_vs_rand}). Results are averaged across budgets within each question,
        then equally across questions.
    }
    \vspace{-1.3em}
    \label{tab:judge-uid-vs-sc}
\end{table}
\section{Training with Auxiliary Trajectory Signals}
\label{sec:uid-training}

We mainly investigate auxiliary trajectory signals using post-training methods for reasoning models in online TTRL~\citep{zuo2026ttrl}, where scores refine majority-vote feedback without reference answers. We also extend our setup to offline training, where scores define DPO~\citep{rafailov2023direct} preferences and SFT weights among correct responses. Therefore, majority agreement is a proxy for correctness during online setting, whereas offline verification fixes final-answer correctness explicitly. These settings let us examine the usefulness of auxiliary preferences under different forms of supervision and optimization. Together, these settings test whether trajectory-level distinctions provide useful supervision beyond answer-level feedback.

\subsection{Auxiliary Trajectory Scoring}
\label{subsec:aux-scores}

For each response $y_i$ to a prompt $x$, we compute an auxiliary score $q_i$ from token-level predictive statistics. We consider UID-Conj., which combine global variation and local smoothness of the entropy trajectory, and SC, summarizing predictive confidence. All three scores use renormalized top-25 predictive distributions, with larger scores indicating stronger preference under the corresponding rule. Appendix~\ref{app:trajectory-statistics} details the shared token-statistic extraction and trajectory construction.

\paragraph{Online setting.}
We compute detached token statistics from the old policy along
each sampled response. For UID, we partition the entropy
trajectory into 32 relative-position bins and normalize the bin
entropies by their mean magnitude. We measure global variation
using population variance, $G_i^{\mathrm{on}}$, and local
smoothness using the negative mean absolute difference between
adjacent bins, $S_i^{\mathrm{on}}$. Each feature is mapped to an
empirical midrank quantile within the prompt's eligible response
group:
\begin{equation}
    u_i^{\mathrm{on}}=Q_x(G_i^{\mathrm{on}}),
    \qquad
    v_i^{\mathrm{on}}=Q_x(S_i^{\mathrm{on}}).
\end{equation}
UID-Conj multiplies these quantiles:
\begin{equation}
    q_i^{\mathrm{Conj,on}}
    =u_i^{\mathrm{on}}v_i^{\mathrm{on}}
    \label{eq:online-uid-scores}
\end{equation}
For SC, we apply the same ranking procedure to its
response-token mean. Ranking is independent of majority
agreement: eligible responses contribute regardless of their
final answers. Section~\ref{subsec:online-aux} describes how
majority agreement subsequently gates the auxiliary reward. Exact features and ranking conventions appear in
Appendix~\ref{app:online-scoring}, with eligibility criteria as well.

\paragraph{Offline setting.}
We score a fixed response pool using the frozen base model.
Offline UID uses a shared robust entropy scale and a Huber-based
smoothness measure, yielding features $G_i^{\mathrm{off}}$ and
$S_i^{\mathrm{off}}$. We transform these features using empirical
CDFs fitted on a separate calibration split without correctness
labels:
\begin{equation}
    u_i^{\mathrm{off}}=F_G(G_i^{\mathrm{off}}),
    \qquad
    v_i^{\mathrm{off}}=F_S(S_i^{\mathrm{off}}).
\end{equation}
UID-Conj. apply the same product aggregation rules as in
Equation~\eqref{eq:online-uid-scores}.
SC is calibrated analogously using its
response-token mean. Thus, online and offline scoring share
the aggregation rules but differ in their scoring policies,
feature definitions, and normalization. The offline scorer
and calibration remain fixed, and the resulting scores define
preferences or example weights among correct responses
(Section~\ref{subsec:offline-aux}).
Appendix~\ref{app:offline-scoring} details the offline features
and calibration, and the eligibility criteria.

\subsection{Majority-Gated Online RL}
\label{subsec:online-aux}

Following TTRL~\citep{zuo2026ttrl}, we derive answer-level
supervision from agreement among sampled responses without
using reference answers. For each prompt $x$, let $\hat a_x$
denote the most frequent extracted answer. We define
\begin{equation}
    \begin{aligned}
        g_i
        &= \mathbf{1}[a(y_i)\simeq\hat a_x], \quad
        \mathcal M_x
        &= \{i:g_i=1\},
    \end{aligned}
\end{equation}
where $\simeq$ denotes agreement under the answer-matching
procedure. Majority agreement provides a pseudo-label rather
than a guarantee of correctness, and auxiliary scores do not
affect the vote. Responses in $\mathcal M_x$ receive the same
binary answer reward, motivating an auxiliary signal that
distinguishes their trajectories.

Using the scores from Section~\ref{subsec:aux-scores}, we
construct the reward
\begin{equation}
    \label{eq:majority-gated-reward}
    \begin{aligned}
        R_i
        &= 2g_i-1+\lambda b_i, \\
        b_i
        &=
        \begin{cases}
            \operatorname{clip}(q_i-\bar q_x,-\delta,\delta),
            & i\in\mathcal M_x,\ |\mathcal M_x|\ge2, \\
            0,
            & \text{otherwise},
        \end{cases}
    \end{aligned}
\end{equation}
where $\bar q_x=|\mathcal M_x|^{-1}
\sum_{j\in\mathcal M_x}q_j$ is computed when
$|\mathcal M_x|\ge2$.
Centering expresses preference relative to other
majority-consistent responses, while gating restricts the
adjustment to that group. We use $\lambda=0.1$ and
$\delta=0.5$, bounding the auxiliary adjustment by $0.05$
and preserving the reward ordering between majority-consistent
and inconsistent responses. Groups with fewer than two
majority-consistent responses receive no auxiliary adjustment.

We compute GRPO advantages by normalizing the combined
rewards within each prompt group and update the policy
using a clipped policy objective with a separate reference
KL penalty. Auxiliary scores and advantages are detached.
The same reward construction applies to UID-Conj. and SC.
Appendix~\ref{app:online-optimization} provides the voting,
advantage, and optimization details. The adjustment preserves answer-reward ordering but can
produce nonzero advantages in unanimous groups, where
$\lambda$ approximately cancels under normalization.
Appendix~\ref{app:aux-advantage-normalization} derives this
effect and discusses implications for the random control.

\subsection{Offline Externsions: Learning from Verified Traces}
\label{subsec:offline-aux}

We extend auxiliary trajectory scoring to offline training
using a fixed response pool scored by the frozen base model.
Both extensions retain complete, nontruncated, correct,
score-eligible responses, allowing auxiliary scores to
differentiate trajectories whose final answers are already
correct.

\paragraph{Auxiliary-Score Preferences for DPO.}
For each prompt $x$, let $\mathcal C_x$ denote its eligible
correct responses. We construct a preference pair by selecting
the highest- and lowest-scoring responses:
\begin{equation}
    y_x^+=\arg\max_{y\in\mathcal C_x}q(x,y),
    \qquad
    y_x^-=\arg\min_{y\in\mathcal C_x}q(x,y).
\end{equation}
Both responses are correct, so the preference reflects
the auxiliary score rather than a difference in answer
correctness. We require a minimum score gap and retain the
common set of qualifying prompts across scoring methods.
The resulting pairs are used with the standard
DPO objective~\citep{rafailov2023direct}.
As a control, we randomly select and orient two distinct
correct responses for each prompt in the same set.
Appendix~\ref{app:offline-training} provides the pair-selection
and initialization details.

\paragraph{Auxiliary-Weighted SFT.}
We retain all eligible correct responses and use auxiliary
scores to weight their contributions to supervised training.
For a prompt with $m_x=|\mathcal C_x|$ responses, we assign
\begin{equation}
    w_i
    =
    m_x
    \frac{\exp(\alpha q_i)}
    {\sum_{y_j\in\mathcal C_x}\exp(\alpha q_j)}.
    \label{eq:aux-sft-weights}
\end{equation}
These weights have mean one within each prompt, preserving
its total example weight while emphasizing higher-scoring
responses. We minimize
\begin{equation}
    \mathcal L_{\mathrm{WSFT}}
    =
    -\frac{1}{N}\sum_{i=1}^{N}
    \frac{w_i}{T_i}
    \sum_{t=1}^{T_i}
    \log\pi_\theta(y_{i,t}\mid x_i,y_{i,<t}),
    \label{eq:aux-weighted-sft}
\end{equation}
where $N$ is the number of training responses and $T_i$ is
the response length. Uniform SFT uses $w_i=1$ on the same
examples and with the same response-level loss normalization.
This comparison isolates the effect of auxiliary weighting
while holding the training responses fixed.
\section{Experiment}
\subsection{Experiment Setup}
\label{subsec:exp-setup}

\begin{table*}[t]
\centering
\small
\setlength{\tabcolsep}{4pt}
\begin{tabular}{@{}lrrrrrr@{}}
\toprule
& \multicolumn{1}{c}{\textbf{In-Domain}}
& \multicolumn{3}{c}{\textbf{Out-of-Domain}}
& & \\
\cmidrule(lr){2-2}\cmidrule(lr){3-5}
Method & AIME2026 & HMMT2026 & MinervaMath & GPQA-D
       & Macro-avg. & $\Delta$ \\
\midrule
\multicolumn{7}{@{}l}{\textbf{Qwen3-1.7B}} \\
\addlinespace[3pt]
Base model    & 30.00 & 9.70 & 25.15 & 38.79 & 25.91 & 0.00 \\
TTRL          & 30.67 & 7.27 & 25.51 & 38.59 & 25.51 & -0.40 \\
Random-TTRL   & 32.67 & 7.88 & 25.88 & 37.58 & 26.00 & +0.09 \\
\rowcolor{lightgray!15}
SC-TTRL       & 34.00 & \textbf{10.91} & 25.74 & 39.49 & 27.54 & +1.63 \\
\rowcolor{lightgray!15}
UID-Conj TTRL & \textbf{35.33} & 9.09 & \textbf{26.25} & \textbf{39.70} & \textbf{27.59} & \textbf{+1.68} \\
\midrule
\multicolumn{7}{@{}l}{\textbf{Qwen3-4B}} \\
\addlinespace[3pt]
Base model    & 55.33 & 11.52 & 31.03 & 56.57 & 38.61 & 0.00 \\
TTRL          & \textbf{60.00} & 10.91 & 30.59 & 56.87 & \textbf{39.59} & \textbf{+0.98} \\
\rowcolor{lightgray!15}
SC-TTRL       & 52.67 & 12.12 & 30.37 & 56.06 & 37.81 & -0.81 \\
\rowcolor{lightgray!15}
UID-Conj TTRL & 55.33 & \textbf{13.94} & \textbf{31.40} & \textbf{57.68} & \textbf{39.59} & \textbf{+0.98} \\
\bottomrule
\end{tabular}
\caption{Online effectiveness on competition-level math benchmarks with known answers.
Mean accuracy of 5 is reported in percent.
Macro-avg.\ is the unweighted mean across benchmarks; $\Delta$ is its
percentage-point difference from the base model. Notably, TTRL adapts to AIME2026 using majority-voting rewards rather than ground-truth outcome rewards, explaining its stronger in-domain gains on Qwen3-4B. In contrast, UID-Conj TTRL improves all three out-of-domain benchmarks, suggesting better generalization beyond the adaptation distribution.}\label{tab:online-effectiveness}
\vspace{-0.5em}
\end{table*}
\begin{table*}[t]
\centering
\small
\setlength{\tabcolsep}{4pt}
\begin{tabular}{@{}lrrrrrr@{}}
\toprule
& \multicolumn{1}{c}{\textbf{In-Domain}}
& \multicolumn{3}{c}{\textbf{Out-of-Domain}}
& & \\
\cmidrule(lr){2-2}\cmidrule(lr){3-5}
Method & AIME2026 & HMMT2026 & MinervaMath & GPQA-D
       & Macro-avg. & $\Delta$ \\
\midrule
\multicolumn{7}{@{}l}{\textbf{DPO}} \\
\addlinespace[3pt]
Base              & 30.00 & 9.70 & 25.15 & 38.79 & 25.91 & 0.00 \\
Random-DPO        & \textbf{38.00} & 8.48 & 26.18 & 38.48 & 27.78 & +1.87 \\
\rowcolor{lightgray!15}SC-DPO
                  & 32.00 & 11.52 & 26.18 & 37.58 & 26.82 & +0.91 \\
\rowcolor{lightgray!15}UID-Conj DPO
                  & 37.33 & \textbf{12.73} & \textbf{26.47} & \textbf{39.60} & \textbf{29.03} & \textbf{+3.12} \\
\midrule
\multicolumn{7}{@{}l}{\textbf{Weighted-SFT}} \\
\addlinespace[3pt]
Base              & 30.00 & 9.70 & 25.15 & 38.79 & 25.91 & 0.00 \\
Uniform-SFT        & 34.67 & 10.91 & \textbf{26.40} & 38.28 & 27.56 & +1.65 \\
\rowcolor{lightgray!15}SC-SFT
                  & 37.33 & 9.70 & 26.03 & 39.49 & 28.14 & +2.23 \\
\rowcolor{lightgray!15}UID-Conj SFT
                  & \textbf{39.33} & \textbf{12.12} & 26.10 & \textbf{40.40} & \textbf{29.49} & \textbf{+3.58} \\
\bottomrule
\end{tabular}
\par\smallskip
\begin{minipage}{0.95\textwidth}
\end{minipage}
\caption{Offline effectiveness on competition-level math benchmarks with known answers.
Mean acc. of 5 is reported in percent; $\Delta$ is the macro-average difference
from the base model in percentage points.}\label{tab:offline-effectiveness}
\vspace{-1.3em}
\end{table*}
\begin{table*}[t]
\centering
\small
\setlength{\tabcolsep}{5pt}
\resizebox{\linewidth}{!}{%
\begin{tabular}{@{}l*{5}{r}|*{7}{r}@{}}
\toprule
& \multicolumn{5}{c|}{\textbf{Out-of-Domain Reasoning}}
& \multicolumn{7}{c}{\textbf{Instruction-Following}} \\
\cmidrule(lr){2-6}
\cmidrule(lr){7-13}
& \multicolumn{2}{c}{CriticBench}
& LabBench2
& &
& \multicolumn{3}{c}{IHEval}
& \multicolumn{2}{c}{IFBench}
& & \\
\cmidrule(lr){2-3}
\cmidrule(lr){4-4}
\cmidrule(lr){7-9}
\cmidrule(lr){10-11}
Method
& (F1) & (Acc)
& (Acc)
& Macro-Avg. & $\Delta$
& Ref. & Align. & Conflict
& Loose & Strict
& Macro-Avg. & $\Delta$ \\
\midrule
Base model
& 62.76 & 67.76
& 41.44
& 57.32 & 0.00
& 72.01 & 69.33 & 11.54
& 21.67 & 17.67
& 38.44 & 0.00 \\
TTRL
& 61.65 & 66.76
& 41.24
& 56.55 & -0.77
& 72.98 & 69.61 & \textbf{12.31}
& 21.67 & 17.67
& 38.85 & +0.40 \\
Random-TTRL
& 63.11 & \textbf{68.00}
& 40.92
& 57.34 & +0.02
& 72.23 & 68.40 & 10.99
& 22.00 & \textbf{18.33}
& 38.39 & -0.05 \\
\rowcolor{lightgray!15}
SC-TTRL
& 62.10 & 67.28
& 41.80
& 57.06 & -0.26
& 73.39 & \textbf{70.03} & 11.73
& \textbf{22.33} & \textbf{18.33}
& \textbf{39.16} & \textbf{+0.72} \\
\rowcolor{lightgray!15}
UID-Conj. TTRL
& \textbf{63.23} & \textbf{68.00}
& \textbf{42.00}
& \textbf{57.74} & \textbf{+0.42}
& \textbf{73.43} & 68.17 & 12.29
& 21.33 & 17.33
& 38.51 & +0.07 \\
\bottomrule
\end{tabular}%
}
\caption{Transfer to critique, scientific reasoning, and instruction following. IFBench uses prompt-level loose and strict accuracy, and IHEval uses task-averaged composite scores. Macro-Avg.\ is the unweighted
mean.
$\Delta$ denotes the percentage-point change in the corresponding
macro-average relative to the base model.}\label{tab:non-veri-transfer}
\vspace{-0.7em}
\end{table*}

\paragraph{Models, training data, and benchmarks.}
We use Qwen3-1.7B as base model, and additionally test on Qwen3-4B as well. Trainings are done for only 2 epochs for RL and 1 epoch for offline training. For online training, we perform test-time adaptation on AIME2026 without reference answers. For offline extensions, we use historical AIME problems gathered until 2025. We evaluate mathematical reasoning on AIME2026, HMMT2026, and MinervaMath, and science reasoning on GPQA's Diamond subset for verifiable tasks, and use CriticBench and LabBench2 for research-level reasoning. We also further evaluate on instruction following tasks to explore the different effects of different auxiliary rewards.

\paragraph{Comparison methods.}
We compare UID-Conj. and SC within each training objective. Online comparisons include the base model, original TTRL, and a random auxiliary-score control. For DPO, we compare auxiliary-score preferences against randomly selected and oriented correct vs. correct pairs. For weighted SFT, uniform weighting provides the control on the same correct response pool.

\paragraph{Evaluation.}
For verifiable mathematical benchmarks, we report mean accuracy across five sampled responses per question. For CriticBench, we report verdict accuracy and incorrect-class F1, which assess verdict correctness rather than critique quality. For IFBench, we report prompt-level loose accuracy using the official verifiers. For IHEval, we use the official task-specific metrics and aggregation procedure, reporting results separately for reference, aligned, and conflicting instruction settings. We document dataset sizes, subset selection, decoding configurations, and training hyperparameters in Appendix~\ref{app:exp-setups}. Our experiments address three research questions:
\begin{enumerate}
    \vspace{-0.2em}
    \item \textbf{Online effectiveness:}
    Do auxiliary trajectory scores improve majority-gated
    test-time adaptation and its transfer to other
    mathematical benchmarks? (Table~\ref{tab:online-effectiveness})

    \item \textbf{Offline effectiveness:}
    Do auxiliary scores provide useful preferences and
    example weights among correct responses
    for DPO and SFT? (Table~\ref{tab:offline-effectiveness})

    \item \textbf{Transfer beyond verifiable tasks:}
    Do models trained with auxiliary signals
    improve error detection on critique-verdict prediction and evidence-conditioned scientific reasoning without task-specific training? (Table~\ref{tab:non-veri-transfer})
    \vspace{-0.2em}
\end{enumerate}
\subsection{Results}

\paragraph{Online effectiveness.}
\label{sec:online-effective}
As seen in Table~\ref{tab:online-effectiveness}, augmenting TTRL with auxiliary rewards improves online adaptation in general. Both SC and UID-Conj. outperform unmodified TTRL on all four benchmarks, increasing macro-average accuracy from 25.51\% to 27.54\% and 27.59\%, respectively. UID-Conj. achieves the highest accuracy on AIME2026, MinervaMath, and GPQA-D, while SC performs best on HMMT2026. Qwen3-4B results reveal a trade-off between target-benchmark gains and downstream transfer. Unmodified TTRL increases AIME2026 accuracy from 55.33\% to 60.00\% but reduces performance on HMMT2026 and MinervaMath, a pattern consistent with over-specialization to the adaptation set. UID-Conj. instead preserves base-model AIME2026 accuracy while improving HMMT2026 and MinervaMath to 13.94\% and 31.40\%. Thus, although auxiliary rewards do not uniformly improve target-benchmark accuracy, UID-Conj. can mitigate the transfer degradation associated with test-time adaptation.

\paragraph{Offline extension.}
\label{sec:offline-effective}
UID-Conj. improves aggregate performance under both offline training objectives (Table~\ref{tab:offline-effectiveness}), achieving macro-average accuracies of 29.03\% with DPO and 29.49\% with weighted SFT, exceeding their respective controls by 1.25 and 1.93 percentage points. UID-Conj. DPO performs best on HMMT2026, MinervaMath, and GPQA-D, although Random-DPO slightly outperforms it on AIME2026 (38.00\% versus 37.33\%). Similarly, UID-Conj. SFT leads on three benchmarks, while uniform SFT retains the highest MinervaMath accuracy (26.40\% versus 26.10\%). SC yields less consistent gains, outperforming uniform SFT in aggregate but falling below Random-DPO. These results show that auxiliary preferences among correct responses can improve training outcomes, while the relative effectiveness of SC and UID-Conj differs from their ordering in the trace-quality diagnostic. This contrast is observed across separate diagnostic and training cohorts; it does not establish that lower judge-rated traces are individually more useful for learning. Explanations are further discussed in Appendix~\ref{app:auxiliary-discussion}.

\paragraph{Evaluation on task transfer.}
\label{sec:transfer-non-veri}
UID-Conj. transfers gains from mathematical training to critique-verdict prediction without task-specific supervision (Table~\ref{tab:non-veri-transfer}). It achieves the highest CriticBench F1 and accuracy at 63.23\% and 68.00\%, respectively, and improves both metrics over the base model. Their average reaches 65.62\%, exceeding the base model by 0.36 percentage points, unmodified TTRL by 1.41 points, and SC by 0.93 points. This result extends UID-Conj's training utility beyond math answer generation, despite SC's stronger trace-quality results in Section~\ref{subsec:preliminary-trace-quality}. Transfer remains task-dependent, however: SC achieves a higher instruction-following aggregate than UID-Conj (39.16\% versus 38.51\%). SC averages $D_{\mathrm{KL}}(U_{25}\|p_t)$ over response tokens, rewarding departure from uniformity in the predictive distribution and thereby favoring answers that they are more familiar with. UID-Conj. instead captures global entropy variation and local smoothness. These distinct preferences are consistent with the  for SC’s higher instruction-following aggregate and UID-Conj’s stronger critique-verdict results in our experiments.

\section{Further Analyses}
\paragraph{Uncertainty Analysis.}
\label{sec:uncertainty-analysis}
\begin{table}[t]
  \centering
  \small
  \setlength{\tabcolsep}{3pt}
  \renewcommand{\arraystretch}{1.08}

  \begin{tabularx}{\columnwidth}{
    @{}
    >{\raggedright\arraybackslash}X
    >{\centering\arraybackslash}p{0.10\columnwidth}
    >{\centering\arraybackslash}p{0.27\columnwidth}
    @{}
  }
    \toprule
    Comparison & $\Delta$ & 95\% CI \\
    \midrule

    \multicolumn{3}{@{}l@{}}{\textit{Online}} \\
    UID-Conj vs.\ TTRL
      & $+1.22$ & $[-0.25,\ 2.85]$ \\
    UID-Conj vs.\ Rand.-TTRL
      & $+1.23$ & $[-0.06,\ 2.54]$ \\
    \addlinespace[4pt]

    \multicolumn{3}{@{}l@{}}{\textit{Offline}} \\
    UID-Conj DPO vs.\ Rand.-DPO
      & $+1.25$ & $[-0.90,\ 3.47]$ \\
    UID-Conj SFT vs.\ Uni.-SFT
      & $+1.93$ & $[0.44,\ 3.44]$ \\
    \addlinespace[4pt]

    \multicolumn{3}{@{}l@{}}{\textit{CriticBench: Acc.}} \\
    UID-Conj vs.\ TTRL
      & $+1.24$ & $[0.20,\ 2.32]$ \\
    UID-Conj vs.\ Rand.-TTRL
      & $0.00$ & $[-1.00,\ 1.00]$ \\
    \addlinespace[4pt]

    \multicolumn{3}{@{}l@{}}{\textit{CriticBench: F1}} \\
    UID-Conj vs.\ TTRL
      & $+1.58$ & $[0.18,\ 3.01]$ \\
    UID-Conj vs.\ Rand.-TTRL
      & $+0.12$ & $[-1.19,\ 1.50]$ \\
    \addlinespace[4pt]

    \multicolumn{3}{@{}l@{}}{\textit{LabBench2: score}} \\
    UID-Conj vs.\ TTRL
      & $+0.76$ & $[-0.80,\ 2.32]$ \\

    \bottomrule
  \end{tabularx}

  \caption{Paired differences for Qwen3-1.7B in percentage
    points (positive favors UID-Conj), with 95\% paired
    prompt-cluster bootstrap CIs from
    100,000 replicates.}
  \label{tab:bootstrap_ci}
\end{table}

We report 95\% paired prompt-cluster bootstrap confidence
intervals for performance differences, retaining all five
responses within each resampled prompt. The primary
multi-benchmark estimates weight benchmarks equally.
The online gains of UID-Conj over TTRL and Random-TTRL have
intervals that include zero, as does the offline DPO comparison.
The offline weighted-SFT comparison has a positive interval,
as do the CriticBench accuracy and F1 comparisons against TTRL.
However, the CriticBench comparisons against Random-TTRL and
the LabBench2 comparison against TTRL include zero. Appendix~\ref{sec:paired-bootstrap-procedure} describes resampling and reports a secondary prompt-weighted analysis.

\paragraph{Training with a different dataset.}
\begin{table}[t]
\centering
\small
\setlength{\tabcolsep}{6pt}
\resizebox{\columnwidth}{!}{%
\begin{tabular}{lcccccc}
\toprule
& \multicolumn{4}{c}{Verifiable}
& \multicolumn{2}{c}{Transfer} \\
\cmidrule(lr){2-5} \cmidrule(lr){6-7}
Method & AIME & HMMT & MM & GPQA-D & CriticBench & LabBench2 \\
\midrule
Base Model & 30.00 & 9.70 & 25.15 & 38.79 & 62.76 & 41.44 \\
TTRL   & 35.33 & 13.33 & 25.07 & 40.71 & 62.46 & 41.20 \\
Random-TTRL & 38.67 & 10.91 & 25.66 & 38.48 & \textbf{63.86} & 41.52 \\
\rowcolor{lightgray!15}
SC-TTRL    & \textbf{41.33} & \textbf{13.94} & \textbf{26.25} & 39.80 & 63.02 & 41.08 \\
\rowcolor{lightgray!15}
UID-Conj. TTRL    & 37.33 & \textbf{13.94} & 26.10 & \textbf{41.21} & \underline{63.19} & \textbf{42.00} \\
\bottomrule
\end{tabular}%
}
\caption{Inference on models trained on S1K datasets.}
\label{tab:s1k-results}
\end{table}
To examine whether the benefits of auxiliary rewards persist across training datasets, we additionally train Qwen3-1.7B on the S1K dataset and evaluate performance on four verifiable reasoning benchmarks and CriticBench and LabBench2 (Table~\ref{tab:s1k-results}). Both SC and UID-Conj. improve over standard TTRL on most benchmarks, with UID-Conj. consistently outperforming TTRL across five. These results suggest that the benefits of auxiliary rewards are not specific to the original training dataset and extend to a different training setting.

\section{Conclusion}
Correct answers do not exhaust the learning signal in model-generated reasoning. Our results show that auxiliary trajectory preferences can extract additional training value from existing responses under limited supervision, with the clearest evidence coming from UID-Conj-weighted SFT on verifier-correct traces. They also reveal a distinction between judged trace quality and training utility: SC selects higher-rated traces, while UID-Conj yields larger gains in our offline training comparisons and selected transfer tasks. These findings motivate evaluating trajectory preferences by the learning outcomes they support, alongside their ability to identify high-quality reasoning.

\section*{Limitations}

Our experiments focus on Qwen3-1.7B, with additional online evaluation on Qwen3-4B, under limited mathematical supervision. Online AIME2026 results measure adaptation on training problems, while the remaining benchmarks evaluate transfer. Our online comparisons do not fully distinguish the benefit of trajectory-specific preferences from the effect of introducing reward variation into unanimous groups.
To address this, we perform an experiment with a control that permutes the realized auxiliary terms within majority-agreeing groups would more closely isolate their
association with individual trajectories
(Appendix~\ref{app:aux-advantage-normalization}).

In addition, the trace-quality diagnostic currently uses one model judge on verifier-correct responses. Transfer evaluation measures critique-verdict accuracy, evidence-conditioned scientific responses, and programmatically verified instruction following, rather than unrestricted open-ended reasoning. Broader response cohorts and human validation would extend this assessment. In particular, the observed difference between trace-selection quality and training utility motivates developing auxiliary preferences that predict a response’s contribution to learning.

\section*{Broader Impact and Ethical Implications}
Improving compact models from limited training data may make reasoning systems more accessible in settings where larger models or extensive supervision are impractical. Auxiliary signals derived from predictive distributions offer a way to reuse model-generated responses without requiring a separately trained reward model or step-level training annotations. However, generating and scoring multiple trajectories still incurs computational costs, and our experiments do not establish net energy or cost savings.

Self-improvement can also reinforce existing errors: majority agreement may reflect shared misconceptions, while predictive confidence and smooth entropy trajectories do not guarantee factual or logical validity. The observed gains in critique and scientific benchmarks should therefore not be interpreted as evidence of dependable autonomous scientific judgment. Moreover, our ultimate goal is to extend reasoning improvements to tasks where reliable answer-level supervision is unavailable, such as research-level mathematics~\citep{son2026researchmath, glazer2024frontiermath}. Therefore, further investigation should be done on open-ended reasoning with human-expert verification. Applications should evaluate these properties separately, retain independent verification where available, and use expert oversight when incorrect outputs could have consequential effects.






\clearpage
\bibliography{custom}

\appendix
\section{Details of Preliminary Diagnostic}

\subsection{Selector Metric Calculation}
\label{app:selector-metrics}

For each question and model, selectors operate on nested prefixes of
$N\in\{2,4,8\}$ generated responses. Final answers are extracted from the
last \texttt{\textbackslash boxed\{\}} expression and normalized before
answer aggregation. Correctness is evaluated after selection by matching
the normalized answer against the normalized reference. Abstentions,
incomplete responses, and truncated responses count as incorrect.
Reference-answer correctness is not used by the answer-accuracy selectors.

\paragraph{Majority Voting.}
Let $a_i$ denote the normalized answer extracted from response $i$.
Among complete responses with an extractable answer, majority voting
selects
\[
    a^\star
    = \operatorname*{arg\,max}_{a}
      \sum_i \mathbf{1}[a_i=a].
\]
Repeated samples contribute separate votes, even when their response
texts are identical. Ties are resolved by the earliest occurrence in
generation order, and the earliest response supporting the winning
answer is retained as its representative. If no answer can be extracted
from any complete response, the selector abstains.

\paragraph{Self-Certainty.}
We mainly follow the setup used in ~\citep{kang2025scalable}. We rescore each generated response under its frozen generating model, using the raw next-token logits without rollout-temperature scaling. Only response-token positions are scored; special tokens and chat control tokens, including thinking delimiters, are excluded. However, unlike the original paper, which computes SC over the full vocabulary, we compute it over the top-$25$ highest-logit tokens at each position for efficiency, renormalizing their probabilities and using a uniform reference distribution over this restricted support. At each retained position $t$, let $\mathcal{K}_t$ contain the $K=25$ highest-logit tokens and define
\[
    p_{t,j}
    = \frac{\exp(z_{t,j})}
           {\sum_{k\in\mathcal{K}_t}\exp(z_{t,k})},
    \qquad j\in\mathcal{K}_t.
\]
The response-level Self-Certainty score is
\[
\begin{aligned}
    \mathrm{SC}(y)
    &= \frac{1}{T}\sum_{t=1}^{T}
       D_{\mathrm{KL}}\!\left(U_K\,\|\,p_t\right) \\
    &= \frac{1}{T}\sum_{t=1}^{T}
       \left[-\frac{1}{K}\sum_{j\in\mathcal{K}_t}\log p_{t,j}
             -\log K\right].
\end{aligned}
\]
where $T$ is the number of retained response tokens and $U_K$ is uniform
over the top-$K$ support. Thus, the implemented score uses a renormalized
top-25 distribution rather than the full vocabulary.

Scores are mapped to calibration empirical-CDF midranks. For calibration
values $\{x_m\}_{m=1}^{M}$, we use
\[
    \widehat F(x)
    = \frac{1}{M}\sum_{m=1}^{M}
      \left(\mathbf{1}[x_m<x]
            +\frac{1}{2}\mathbf{1}[x_m=x]\right).
\]
The SC selector ranks responses by
$q_{\mathrm{SC}}(y)=\widehat F_{\mathrm{SC}}(\mathrm{SC}(y))$.
For comparability with UID, eligible candidates must be complete,
non-truncated, and contain at least 32 scored tokens. The direct selector
returns the highest-ranked eligible response, breaking score ties by
generation order.

For the Borda-added variant, eligible responses are ordered by decreasing
$q_{\mathrm{SC}}$. A response at one-based rank $r$ among $m$ eligible
responses receives weight $\sqrt{m-r+1}$. We sum these weights by
normalized answer and select the answer with the largest total. Responses
without extractable answers contribute no answer weight, although they
remain in the ranking. Tied answer totals are resolved by their first
occurrence in ranked order; the highest-ranked response supporting the
winning answer is retained.

\paragraph{UID-Conj.}
UID-Conj uses the token entropies of the same top-25 distributions,
\[
    H_t=-\sum_{j\in\mathcal{K}_t}p_{t,j}\log p_{t,j}.
\]
We partition the $T$ retained positions into $B=32$ contiguous
relative-position bins, with boundaries
$b_k=\lfloor kT/B\rfloor$, and average entropy within each bin:
\[
    h_k=\frac{1}{b_k-b_{k-1}}
        \sum_{t=b_{k-1}+1}^{b_k}H_t,
    \qquad k=1,\ldots,B.
\]
A model-specific robust scale is fitted using the pooled bin entropies
$\mathcal{H}_{\mathrm{cal}}$ of eligible calibration responses:
\[
\begin{aligned}
    s = \max\!\Bigl\{&
      1.4826\,
      \operatorname{median}_{h\in\mathcal{H}_{\mathrm{cal}}}
      \left|h-\operatorname{median}(\mathcal{H}_{\mathrm{cal}})\right|, \\
      &10^{-3}
    \Bigr\}.
\end{aligned}
\]
The global-variation and local-smoothness features are
\[
\begin{aligned}
    G(y) &= \frac{1}{B}\sum_{k=1}^{B}
             \left(\frac{h_k-\bar h}{s}\right)^2, \\
    S(y) &= -\frac{1}{B-1}\sum_{k=1}^{B-1}
             \rho\!\left(\frac{h_{k+1}-h_k}{s}\right).
\end{aligned}
\]
where $\bar h=B^{-1}\sum_k h_k$ and
\[
    \rho(u)=
    \begin{cases}
        \frac{1}{2}u^2, & |u|\leq 1,\\
        |u|-\frac{1}{2}, & |u|>1
    \end{cases}
\]
is the unit-threshold Huber loss. Larger $G$ corresponds to greater
global entropy variation, whereas larger $S$ corresponds to smoother
local changes. In particular, the implemented global feature does not
reward globally constant entropy.

Using separate calibration midrank functions for these features, the
conjunction score is
\[
    q_{\mathrm{UID\text{-}Conj}}(y)
    = \widehat F_G(G(y))\,\widehat F_S(S(y)).
\]
The robust scale and empirical distributions are fitted only on
calibration responses, without correctness or judge labels, and are
subsequently frozen. Calibration/evaluation prompt overlap is checked.
Direct UID-Conj selection and its Borda-added variant use the same
eligibility, tie-breaking, and square-root rank-weighting rules as SC,
with $q_{\mathrm{UID\text{-}Conj}}$ replacing $q_{\mathrm{SC}}$.

\subsection{Details of Trace Quality Assessment}
\label{app:trace-quality}

The trace-quality diagnostic asks whether auxiliary scores distinguish
reasoning quality among responses with correct final answers. It uses
the saved responses and selector outputs from the answer-correctness
experiment. The reported analysis considers Qwen3-1.7B and Qwen3-4B on
MinervaMath, HMMT 2026, and AIME 2026; MinervaMath's calibration and
development questions are combined. Consequently, this analysis is conditional on the frozen
response pools and is not an exclusively held-out evaluation.

\paragraph{Stage1.}
We construct comparison cohorts and blinded assessment requests.
For each model, question, and rollout budget $N\in\{2,4,8\}$, a pool
qualifies only if it contains at least two distinct response texts that
are complete, non-truncated, UID-eligible, and verifier-correct.
The main, correct-only diagnostic selects the highest-ranked SC and
UID-Conj responses within these correct candidates. Its random-correct
baseline is one uniform draw over distinct eligible correct response
texts. The same baseline draw is reused across contrasts.

We construct three contrasts: UID-Conj versus random correct, SC versus
random correct, and UID-Conj versus SC. A separate supporting study uses
the exact SC and UID-Conj Borda representatives selected from the full
candidate pool. Pool eligibility alone does not ensure that these Borda
representatives are correct; their correctness is checked during the
final analysis.

Each judge request contains only the problem and one full response,
together with the fixed rubric and output schema. Reference answers,
model and selector identities, competing responses, and comparison
context are withheld. Identical assessment inputs are deduplicated and
reuse one assessment across comparisons. Before submission, the pipeline
counts input tokens, checks request-size limits and the configured
maximum-cost budget, and freezes request identities and provenance.

\paragraph{Stage2.}
Each unique response is assessed independently using
\texttt{gpt-6-luna}~\citep{openai_gpt6_luna} with high reasoning effort and a maximum output budget
of 12,000 tokens. The recorded run uses concurrent standard Responses
requests. The judge returns a structured assessment containing an
integer score from 1 to 5, supporting evidence, and an explanation for
each of four equally weighted criteria:
\begin{itemize}
    \item[$\circ$] \textbf{Validity:} whether the written argument is logically
    and mathematically sound.
    \item[$\circ$] \textbf{Sufficiency:} whether the response supplies the
    justification necessary to support its conclusion.
    \item[$\circ$] \textbf{Consistency:} whether the reasoning contains
    unresolved internal contradictions.
    \item[$\circ$] \textbf{Relevance/non-redundancy:} whether the reasoning serves
    the solution without unnecessary detours or repetition.
\end{itemize}
Scores range from severe deficiencies at 1 to no supported deficiency
at 5. The rubric instructs the judge to evaluate the reasoning actually
written, without repairing it or inferring correctness from confident
style. Routine calculations need not be expanded; explicitly corrected
false starts are not treated as contradictions; and length alone is not
a defect. Issues are documented in the explanations without additional
numerical penalties. If a criterion cannot be assessed reliably, the
judge returns \texttt{null} with an explanation rather than assigning
a low score.

\paragraph{Stage3.}
We validate request coverage and assessment schemas, then derive
comparisons from the independent scores. For an assessable response $y$,
the total score is
\[
    Q(y)=V(y)+F(y)+C(y)+R(y)\in[4,20],
\]
where $V,F,C,R$ denote validity, sufficiency, consistency, and
relevance/non-redundancy, respectively. We additionally compute the
substantive subtotal
\[
    Q_{\mathrm{sub}}(y)=V(y)+F(y)+C(y)\in[3,15]
\]
to distinguish substantive reasoning differences from relevance or
repetition effects.

A response wins when its total exceeds that of its comparator.
Equal totals for different responses are recorded as score ties;
identical assessment inputs are retained as a separate identical-response
tie category. These outcomes are derived from independently assigned
scores, rather than direct pairwise choices by the judge. Missing,
invalid, or partially unassessable assessments are excluded from scored
estimates and reported separately, never counted as losses.

For the primary analysis, all three contrasts must be assessable and
both responses in every contrast must be verifier-correct within each
study/model/question/budget combination. This yields a common
UID-Conj/SC/random cohort within each budget. Correct-only ranking is
the main analysis; full-pool Borda with subsequent both-correct filtering
is reported separately because its qualifying cohort can differ.

We report win, loss, and both tie rates; net preference
$P(\mathrm{win})-P(\mathrm{loss})$; and paired differences in total,
substantive, and individual-criterion scores. Pooled estimates first
average over the available eligible budgets within each question and
then weight questions equally. Repeated budgets therefore do not count
as independent questions, and benchmarks contribute according to their
eligible question counts. Numeric uncertainty estimates use 4,000
question-level bootstrap resamples and 95\% percentile
intervals. The figures and displayed tables show point estimates.
These comparisons describe the saved judge run and fixed random
baseline; they do not establish agreement across judges or benefits
from using the scores as training rewards.

\section{Details of Online-and-Offline Training}
\subsection{Auxiliary Trajectory Scoring}
\label{app:trajectory-scoring}

\paragraph{Token statistics.}
\label{app:trajectory-statistics}
At each valid response position $t$, we renormalize the
scoring model's top-$K$ logits, with $K=25$, to obtain
$\tilde p_t$. We compute entropy and Self-Certainty as
\begin{equation}
    \begin{aligned}
        H_t
        &= -\sum_{v=1}^{K}
        \tilde p_t(v)\log\tilde p_t(v), \\
        C_t
        &= -\frac{1}{K}\sum_{v=1}^{K}
        \log\tilde p_t(v)-\log K.
    \end{aligned}
\end{equation}
These statistics use logits before sampling-temperature
scaling and exclude padding and special tokens.
Self-Certainty is therefore a top-25 approximation.
We average $C_t$ over valid response tokens and partition
the entropy trajectory into $B=32$ relative-position bins,
whose mean entropies are denoted by $h_{i,b}$.

\paragraph{Online scoring.}
\label{app:online-scoring}
Let
\begin{equation}
    \begin{aligned}
        \bar h_i
        &= \frac{1}{B}\sum_{b=1}^{B} h_{i,b}, \\
        \tilde h_{i,b}
        &= \frac{h_{i,b}}{\max(|\bar h_i|,10^{-8})}.
    \end{aligned}
\end{equation}
The online UID features are
\begin{equation}
    \begin{aligned}
        G_i^{\mathrm{on}}
        &= \operatorname{Var}_{b}(\tilde h_{i,b}), \\
        S_i^{\mathrm{on}}
        &= -\frac{1}{B-1}\sum_{b=1}^{B-1}
        \left|\tilde h_{i,b+1}-\tilde h_{i,b}\right|,
    \end{aligned}
\end{equation}
where variance uses the population convention.
For the eligible response pool $\mathcal E_x$, the midrank
transform is
\begin{equation}
    \begin{aligned}
        Q_x(z)
        &= \frac{1}{|\mathcal E_x|}
        \sum_{j\in\mathcal E_x}
        \left(
            \mathbf{1}[z_j<z]
            +\tfrac12\mathbf{1}[z_j=z]
        \right).
    \end{aligned}
\end{equation}
UID-Conjunction combine the feature quantiles by multiplication. Self-Certainty applies $Q_x$ to the response-token mean of $C_t$. All statistics are detached.

\paragraph{Offline scoring.}
\label{app:offline-scoring}
We fit a shared robust scale
\begin{equation}
    \begin{aligned}
        s
        &= \max\left(
            1.4826\,\operatorname{MAD}(\mathcal H),
            10^{-3}
        \right),
    \end{aligned}
\end{equation}
where $\mathcal H$ contains the bin entropies of eligible
calibration responses. The offline features are
\begin{equation}
    \begin{aligned}
        G_i^{\mathrm{off}}
        &= \frac{1}{B}\sum_{b=1}^{B}
        \left(\frac{h_{i,b}-\bar h_i}{s}\right)^2, \\
        S_i^{\mathrm{off}}
        &= -\frac{1}{B-1}\sum_{b=1}^{B-1}
        \rho_1\left(\frac{h_{i,b+1}-h_{i,b}}{s}\right),
    \end{aligned}
\end{equation}
where
\begin{equation}
    \rho_1(d)=
    \begin{cases}
        \dfrac{d^2}{2}, & |d|\le 1, \\[4pt]
        |d|-\dfrac{1}{2}, & \text{otherwise}.
    \end{cases}
\end{equation}
Each feature and response-level Self-Certainty is mapped
through an empirical CDF using the same midpoint convention
as $Q_x$. Calibration uses a shared UID-eligible subset,
without correctness filtering, from prompts disjoint from
training and development. The scoring model, scale, and
CDFs remain fixed.

\paragraph{Eligibility.}
\label{app:scoring-eligibility}
Both UID variants require at least 32 valid scoring tokens;
online Self-Certainty requires at least one.
Online ineligible responses receive score zero and are
excluded from ranking, but still enter the gate-positive
score mean if they agree with the majority answer.
Offline calibration requires complete, nontruncated,
UID-eligible responses; offline training additionally
requires verifier correctness and finite scores for all
compared methods.

\subsection{Majority-Gated Online Optimization}
\label{app:online-optimization}

We follow TTRL's answer extraction and matching procedure.
The most frequent extracted answer supplies the pseudo-label,
with voting ties resolved by first occurrence. Auxiliary
scores do not influence this selection.

For each prompt group of $n>1$ responses, we normalize the
combined rewards from Equation~\eqref{eq:majority-gated-reward}:
\begin{equation}
    A_i=
    \frac{R_i-\bar R_x}
    {\operatorname{Std}(R_1,\ldots,R_n)+10^{-6}},
    \quad
    \bar R_x=\frac1n\sum_{j=1}^{n}R_j,
\end{equation}
where $\operatorname{Std}$ denotes the sample standard
deviation. The detached sequence advantage is shared across
response tokens. The original TTRL control uses binary
$0/1$ rewards, while auxiliary arms use the $-1/+1$ answer
reward plus the gated adjustment.

We retain the implementation's clipped GRPO policy update,
with clipping threshold $0.2$, dual-clipping parameter $3$,
and a separate reference KL penalty with coefficient
$10^{-3}$. KL regularization does not enter the rewards
used for advantage normalization.

\subsubsection{Effect of Auxiliary Rewards under Advantage Normalization}
\label{app:aux-advantage-normalization}

The bounded auxiliary adjustment preserves the ordering of
answer-level rewards, but does not necessarily imply a small
change in the normalized advantages. To distinguish these
properties, write the online reward as
\[
R_i = 2g_i - 1 + \lambda b_i,
\]
where $g_i$ indicates majority agreement and $b_i$ is the
majority-gated, centered, and clipped auxiliary term.
With $\lambda=0.1$ and $|b_i|\leq 0.5$, every
majority-agreeing response receives a reward of at least
$0.95$, whereas every disagreeing response receives $-1$.
Thus, the auxiliary term cannot reverse the reward ordering
between these two groups. However, GRPO normalizes the
combined rewards within each prompt group, so this bound
does not directly bound the resulting change in advantages.

\paragraph{Unanimous groups.}
Consider a group of $n>1$ responses in which every response
agrees with the majority answer, so that $g_i=1$ for all $i$.
Define
\[
\bar b = \frac{1}{n}\sum_{i=1}^{n} b_i,
\qquad
s_b = \operatorname{Std}(b_1,\ldots,b_n),
\]
using the same sample-standard-deviation convention as the
training implementation. The mean reward is
$1+\lambda\bar b$, and, for $\lambda>0$, its standard
deviation is $\lambda s_b$. Consequently, the normalized
advantage is exactly
\begin{equation}
A_i
=
\frac{\lambda(b_i-\bar b)}
     {\lambda s_b+\epsilon}
=
\frac{b_i-\bar b}
     {s_b+\epsilon/\lambda},
\label{eq:unanimous-aux-advantage}
\end{equation}
where $\epsilon=10^{-6}$ is the numerical stabilizer.
Here, $\bar b$ is retained explicitly because clipping
after centering need not preserve a zero mean.

When $\lambda s_b \gg \epsilon$, the auxiliary coefficient
approximately cancels:
\begin{equation}
A_i \approx
\frac{b_i-\bar b}{s_b}.
\label{eq:unanimous-aux-advantage-approx}
\end{equation}
Thus, nonconstant auxiliary rewards can yield normalized
advantages with approximately unit sample standard deviation,
despite their small absolute reward scale. Under original
TTRL, a unanimous group instead has constant answer rewards
and therefore zero reward-based advantages. Auxiliary
rewards can consequently activate a policy-gradient signal
for groups that provide no such signal under the
answer-only objective. If all auxiliary terms are identical,
the advantages remain zero. Unanimity also does not imply
correctness: all sampled responses may agree on an
incorrect answer.

\paragraph{Mixed-agreement groups.}
When both agreeing and disagreeing responses are present,
the answer reward itself varies. Let
$\bar g=n^{-1}\sum_i g_i$. The centered reward becomes
\[
R_i-\bar R
=
2(g_i-\bar g)+\lambda(b_i-\bar b).
\]
In this case, $\lambda$ generally does not cancel, because
the denominator depends jointly on answer-reward variation
and auxiliary variation. The influence of the auxiliary
term therefore depends on the group's agreement pattern
and score distribution. Reward-order preservation alone
does not characterize this influence. Moreover, normalized
advantages do not directly determine parameter-update
magnitudes, which also depend on policy gradients,
importance ratios, clipping, and the separate KL penalty.

\paragraph{Interpretation of controls and further diagnostics.}
\label{app:perm-control}
The random auxiliary-score control tests whether adding
uninformative within-group reward variation can account for
the observed gains. However, independent uniform scores do
not exactly preserve the realized score distribution of
UID-Conj, including its ties and dispersion.

A more closely matched control would first compute the
auxiliary terms normally and then randomly permute them
among majority-agreeing responses within each prompt group.
For a permutation $\pi$ of the majority-agreeing set
$\mathcal{M}_x$, define
\[
b_i^{\mathrm{perm}} =
\begin{cases}
b_{\pi(i)}, & i\in\mathcal{M}_x,\\
0, & i\notin\mathcal{M}_x.
\end{cases}
\]
This preserves the group's reward multiset, mean, standard
deviation, and normalized-advantage distribution, while
randomizing the association between auxiliary preferences
and response trajectories.

\subsection{Offline Training}
\label{app:offline-training}

\paragraph{Preference construction and initialization.}
We deduplicate identical response strings within each prompt
and select one highest-versus-lowest score pair, requiring
a score gap of at least $0.05$. We retain only prompts that
satisfy this condition for every auxiliary scoring method.
The random control samples two distinct correct responses
from each retained prompt and assigns their orientation
randomly. All DPO arms initialize from a shared SFT checkpoint
trained on the union of responses in the training preference
pairs; development responses are excluded. This checkpoint
also supplies the frozen reference policy. We use DPO
coefficient $\beta=0.1$.

\paragraph{SFT weighting and matched data.}
We use $\alpha=2$ in Equation~\eqref{eq:aux-sft-weights}.
Uniform, Self-Certainty, and UID-Conjunction SFT
use the same eligible responses. Prompt and padding tokens
are masked, and weighted response-mean losses are averaged
over examples across gradient accumulation and data-parallel
workers. Weights are not renormalized within microbatches.
Data matching applies within each objective: DPO shares
a prompt set across arms, while SFT shares a response pool.
The DPO and SFT training subsets need not coincide.

\section{Details of Experiment Setup}\label{app:exp-setups}

\subsection{Details of the Training Dataset}

\paragraph{Online setup.}
Our current online experiments use test-time reinforcement learning
(TTRL), starting from Qwen3-1.7B. For AIME2026, all 30 problems are
used for adaptation and subsequent evaluation. The Math500 configuration
similarly uses all 500 problems for both stages. Thus, these experiments
measure adaptation on the target benchmark rather than generalization
to an unseen test set. Reference answers are retained for evaluation
and diagnostics; training rewards are computed from agreement with
majority-vote pseudo-labels obtained from model-generated responses.

For each prompt, the default configuration generates eight responses
for majority voting and uses all eight for training. We compare
original TTRL, a random-preference control, Self-Certainty (SC), and
UID-based preferences. The auxiliary preferences do not determine the
majority-vote answer. Instead, they modify rewards among responses
that agree with that answer.

\paragraph{Offline setup.}
Offline data are constructed from a shared response pool generated by
Qwen3-1.7B on historical AIME problems. After exact deduplication of
prompt message lists, the current pool contains 882 prompts, partitioned into 88 calibration prompts, 88 development prompts, and
706 training prompts. These partitions are disjoint at the prompt
level. Eight candidate responses are generated per prompt with
temperature 0.7, top-$p$ 0.95, a maximum response length of 16,384
tokens, and a maximum total sequence length of 18,432 tokens.

Calibration uses only the calibration partition and does not use
correctness labels. SC and UID scores are computed under the frozen
base checkpoint using renormalized top-25 token distributions.
Consequently, the implemented SC score is a top-25 approximation
rather than full-vocabulary Self-Certainty. UID-Mix averages the
calibrated global and local component scores, whereas UID-Conjunction
multiplies them.

Training examples must be verifier-correct, complete, non-truncated,
and eligible for UID scoring. Duplicate response strings within a
prompt are removed. The weighted-SFT datasets contain 2,967 training
responses and 355 development responses per method, with identical
examples across methods. The uniform baseline assigns weight one.
For a prompt with $m$ eligible responses, the other methods assign
\[
w_i
=
m\frac{\exp(2q_i)}{\sum_{j=1}^{m}\exp(2q_j)},
\]
where $q_i$ is the corresponding calibrated preference score. This
normalization preserves a mean weight of one within each prompt.
The implementation labels the uniform baseline as Random-SFT.

For DPO, each score-based method selects the highest- and lowest-scoring
eligible responses for a prompt. Both responses are verifier-correct.
We retain only the common set of prompts for which every score-based
method has a preference gap of at least 0.05. This yields 413 training
pairs and 53 development pairs per method. Random-DPO samples and
randomly orients two eligible responses on the same prompt set.

All DPO methods share an SFT initialization constructed from the
deduplicated union of the chosen and rejected responses across the
four DPO datasets. This union contains 1,875 unique training responses
and 226 development responses. The training file repeats 13 responses
to obtain 1,888 rows, a multiple of the global SFT batch size. The
resulting merged SFT checkpoint initializes both the DPO policy and
its frozen reference model.

\subsection{Details of the Benchmark Dataset}

\paragraph{Verifiable Tasks (Table~\ref{tab:verifiable-composition})}
\begin{table*}[t]
\centering

\small
\begin{tabular}{@{}lrlp{6.5cm}@{}}
\toprule
Benchmark & Questions & Coverage & Experimental role \\
\midrule
\multicolumn{4}{@{}l}{\textbf{In-Domain}} \\
\midrule
AIME2026    & 30  & Full snapshot &
Post-training evaluation; also used for AIME test-time adaptation for TTRL experiments. \\
\midrule
\multicolumn{4}{@{}l}{\textbf{Out-of-Domain}} \\
\midrule
HMMT2026    & 33  & Full snapshot &
Post-training evaluation. \\
MinervaMath & 272 & Full snapshot &
Post-training evaluation. \\
GPQA-D     & 198 & Full Diamond subset &
Post-training evaluation of science reasoning. \\
\midrule
Total & 533  \\
\bottomrule
\end{tabular}
\caption{Composition of the verifiable evaluation suite.}
\label{tab:verifiable-composition}
\end{table*}
The verifiable evaluation suite contains 533 questions across four
benchmarks. We generate five sampled responses per question and
report mean sample accuracy using normalized boxed-answer matching.

\begin{itemize}
    \item[$\circ$] \textbf{AIME2026.}\footnote{\url{https://huggingface.co/datasets/math-ai/aime_2026}}
    We use all 30 problems from the pinned
    \texttt{math-ai/aime26} snapshot. Each problem has a reference
    numerical answer. In the AIME online configuration, these same
    problems also serve as the adaptation set.

    \item[$\circ$] \textbf{HMMT2026.}\footnote{\url{https://huggingface.co/datasets/MathArena/hmmt_feb_2026}}
    We use all 33 entries in the pinned
    \texttt{MathArena/hmmt\_feb\_2026} snapshot for post-training
    evaluation. Questions and reference answers are preserved
    from the source.

    \item[$\circ$] \textbf{MinervaMath.}\footnote{\url{https://huggingface.co/datasets/math-ai/minervamath}}
    We use all 272 problems in the pinned
    \texttt{math-ai/minervamath} test snapshot for post-training
    evaluation. This includes the complete source snapshot rather
    than the smaller cohort used in the preliminary analysis.

    \item[$\circ$] \textbf{GPQA-D.}\footnote{\url{https://huggingface.co/datasets/Idavidrein/gpqa}}~\citep{rein2023gpqa}
    We use all 198 questions in the \texttt{gpqa\_diamond}
    configuration of the pinned \texttt{Idavidrein/gpqa} dataset.
    This multiple-choice science benchmark evaluates
    post-training reasoning performance. Prepared questions,
    answer-choice ordering, and reference answers are frozen
    and shared across model configurations.
\end{itemize}

\paragraph{Transfer Tasks (Table~\ref{tab:nonverifiable-composition})}
\begin{table*}[t]
\centering
\small
\begin{tabular}{@{}lrp{5.5cm}p{5cm}@{}}
\toprule
Benchmark & Cases & Selection and composition & Reported metrics \\
\midrule
\multicolumn{4}{@{}l}{\textbf{Out-of-Domain}} \\
\midrule
CriticBench & 500 &
250 correct and 250 incorrect candidate answers; deterministic
selection with coverage across source datasets. &
Verdict accuracy and incorrect-class F1, averaged across five
sample indices. \\
\addlinespace
LabBench2 & 500 &
Deterministic selection of 500 unique question families with
complete frozen text evidence, covering seven task types;
five sampled responses per case. &
Task-specific semantic, numeric, or recall judge scores
in $[0,1]$, averaged across response samples. \\
\midrule
\multicolumn{4}{@{}l}{\textbf{Instruction Following}} \\
\midrule
\addlinespace
IFBench & 300 &
All official single-turn test examples; one greedy response per
example. &
Prompt-level loose accuracy (primary), plus strict/loose prompt
and instruction accuracies. \\
\addlinespace
IHEval & 500 &
Stratified selection across nine tasks and reference, aligned,
and conflict settings; one greedy response per case. &
Official task-specific metrics, aggregated separately for
reference, aligned, and conflict settings. \\
\midrule
Total & 1,800 & & \\
\bottomrule
\end{tabular}
\caption{Composition of the transfer evaluation suite.}
\label{tab:nonverifiable-composition}
\par\smallskip
\begin{minipage}{0.95\textwidth}
\footnotesize
Subset selection uses seed 42. LabBench2 selection is conditioned on text-evidence availability,
rather than representing the full official multimodal benchmark.
Its task composition is sourcequality (132), protocolqa2 (109),
trialqa (80), litqa3 (65), tableqa2-pdf (54), patentqa (46),
and tableqa2 (14).
\end{minipage}
\end{table*}
Ohter transfer evaluation suite contains 1,800 cases across five
benchmarks. Case counts exclude repeated response samples.
CriticBench and LabBench2 use five sampled responses
per case, whereas IFBench and IHEval use one greedy response per case. CriticBench, IFBench and IHEval turns off thinking mode.

\begin{itemize}
    \item[$\circ$] \textbf{CriticBench.}\footnote{\url{https://huggingface.co/datasets/opencompass/CriticBench}}~\citep{lin2024criticbench}
    We use a deterministic 500-example diagnostic subset, comprising
    250 correct and 250 incorrect candidate answers. Selection balances coverage across the available source
    datasets within each correctness class. Inputs contain the
    question and candidate answer, without the reference answer or
    correctness label. We use the official zero-shot critique
    chain-of-thought prompt and request an analysis together with
    a correct/incorrect verdict, with the model's separate thinking
    mode disabled. Evaluation uses the pinned official parser and
    question-only scoring implementation; no generated code is
    executed. We report verdict accuracy and incorrect-class F1,
    computed separately for each of the five sample indices and
    then averaged.

    \item[$\circ$] \textbf{LabBench2.}\footnote{\url{https://huggingface.co/datasets/EdisonScientific/labbench2}}~\citep{laurent2026labbench2} We use an evidence-availability-conditioned subset of 500
unique question families from the train split. Candidate tasks are filtered for complete frozen text evidence and selected, without consulting model outputs or
reference answers. This is a text-based diagnostic subset rather
than a representative sample of the full official multimodal
benchmark.

The subset contains 132 \texttt{sourcequality} cases,
109 \texttt{protocolqa2} cases, 80 \texttt{trialqa} cases,
65 \texttt{litqa3} cases, 54 \texttt{tableqa2-pdf} cases,
46 \texttt{patentqa} cases, and 14 \texttt{tableqa2} cases.
Required source evidence is frozen with the prepared inputs.

Responses are judged using \texttt{gpt-6-luna} through the
OpenAI Responses API, with reasoning effort \texttt{high}, and with
\texttt{max\_output\_tokens=12000}.

We use the upstream judging prompts from LabBench2 repository\footnote{\url{https://github.com/EdisonScientific/labbench2}}.
Specifically, \texttt{tableqa2} and \texttt{tableqa2-pdf}
use {\small\texttt{STRUCTURED\_EVALUATION\_PROMPT\_EXACT\_MATCH}},
which instructs the judge to accept numeric answers with
absolute or relative difference below $10^{-6}$, subject to
explicit formatting requirements. All other selected task
tags use {\small\texttt{STRUCTURED\_EVALUATION\_PROMPT}} for semantic
correctness. The recall prompt is not used for this subset.
Each judging request supplies the complete prepared text
context, reference answer, and model response in a single
user message and requires a structured JSON judgment with
\texttt{rationale} and \texttt{result} fields.
A \texttt{correct} judgment receives 1; \texttt{incorrect}
and \texttt{unsure} receive 0. We average over five response
samples per question and across the 500 questions, retaining
inference failures as zeros.

    \item[$\circ$] \textbf{IFBench.}\footnote{\url{https://huggingface.co/datasets/allenai/IFBench_test}}~\citep{pyatkin2026generalizing}
    We use all 300 examples in the official single-turn test set.
    The set is neither duplicated nor padded to 500. We report
    the official strict and loose prompt-level and instruction-level
    accuracies, with prompt-level loose accuracy as the primary
    metric.

    \item[$\circ$] \textbf{IHEval.}\footnote{\url{https://huggingface.co/datasets/zhihz0535/IHEval}}~\citep{zhang2025iheval}
    We select 500 distinct evaluation cases.
    Selection balances the nine tasks crossed with the reference,
    aligned, and conflict settings, and then distributes examples
    across prompt variants. Cases are identified by their source
    path and original example ID; different variants may share an
    underlying question. The subset covers rule following, task
    execution, safety defense, and tool use. System instructions,
    conversation history, supplied tool calls, and tool outputs
    retain their original roles. Tool outputs are provided by the
    dataset rather than obtained through live tool execution.

    We use the official task-specific metrics, including verb
    extraction F1, translation ROUGE-L, classification accuracy,
    rule-following checks, and safety and tool-use checks.
    Aggregation follows the upstream procedure: combine metrics
    within each variant, average variants within each task and
    setting, and then average the nine tasks separately for the
    reference, aligned, and conflict settings. These are
    mixed-metric aggregates rather than uniform exact-match
    accuracies.
\end{itemize}

Dataset revisions, source hashes, selected case identifiers, and
selection configurations are recorded with the prepared evaluation
inputs. LabBench2 evidence is also frozen and hashed.
IFBench and IHEval use pinned upstream verifier implementations;
they both use
programmatic evaluation rather than an LLM judge. CriticBench uses
the pinned official verdict parser and scorer, while
and LabBench2 use external model judging. Percentage metrics are
obtained by multiplying scores in $[0,1]$ by 100

\subsection{Hyperparameters and GPUs for Training}
\paragraph{GPU and setups.}
We use 4xA100 GPUs for training, and 1xA40 GPU for inference. We use veRL~\citep{sheng2025hybridflow} for TTRL training, and OpenRLHF~\citep{hu2025openrlhf} for DPO an weighted-SFT. We also use vLLM~\citep{kwon2023efficient} with asynchronous continuous batching for efficient inference.  All models were LoRA-trained.
\paragraph{Online setup.}
The default online configuration trains LoRA adapters on Qwen3-1.7B
with rank 16 and scaling parameter 32, targeting all linear layers.
Training uses GRPO with within-prompt standard-deviation normalization,
a policy clipping ratio of 0.2, and one policy-optimization epoch per
rollout batch. The learning rate is $5\times10^{-7}$, with a 3\%
warmup fraction and cosine scheduling. The reference KL-loss
coefficient is $10^{-3}$, and the entropy-loss coefficient is zero.

Training runs for two dataset epochs. The prompt batch size is 30
for AIME2026 and 32 for Math500, retaining the final partial batch.
The configured PPO minibatch size is one prompt, corresponding to
eight sampled responses under the default rollout setting; the
per-GPU microbatch size is one response. The maximum prompt and
response lengths are 2,048 and 16,384 tokens, respectively. Rollouts
use thinking mode, temperature 1.0, top-$p$ 1.0, and no top-$k$
restriction.

For the random, SC, and UID variants, binary majority-agreement
rewards are mapped to $-1/+1$. An auxiliary term
\[
0.1\,\mathrm{clip}
\left(q_i-\overline{q}_{\mathrm{agree}},-0.5,0.5\right)
\]
is added only to majority-consistent responses in groups containing
at least two such responses. Original TTRL retains its binary
$0/1$ reward without this term. Random scores are seeded uniform
draws. SC uses within-prompt midranks of mean response-token
Self-Certainty. Online UID uses 32 equal-position bins and the
product of within-prompt midrank scores for global variation and
local smoothness; it does not use the offline calibration CDFs.

The default configuration uses four GPUs, FSDP, gradient
checkpointing, and CPU offloading of actor parameters, optimizer
state, and reference parameters.
Internal validation uses four sampled responses per prompt with
temperature 0.6 and top-$p$ 0.95, and is distinct from the
post-training five-response evaluation protocol.

\paragraph{Offline setup.}
Offline training uses OpenRLHF with BF16 precision, DeepSpeed
ZeRO-3, FlashAttention-2, sequence packing, and gradient
checkpointing. LoRA adapters have rank 64 and scaling parameter
64 and target the query, key, value, output, gate, up, and down
projections. The maximum sequence length is 18,432 tokens, and
the per-device microbatch size is one.

Weighted SFT and the shared DPO warm-start SFT each run for one
epoch with global batch size 32, learning rate $5\times10^{-5}$,
and a 3\% warmup fraction. Weighted SFT minimizes the mean of
per-response token-averaged negative log-likelihoods multiplied
by the corresponding example weights; prompt and padding tokens
are masked.

DPO runs for one epoch with global batch size 16, learning rate
$10^{-5}$, a 5\% warmup fraction, and preference coefficient
$\beta=0.1$. Full-coverage batching is enabled for the DPO
training pairs. Its reference model is the frozen shared SFT
checkpoint.

Both objectives use AdamW with
$(\beta_1,\beta_2)=(0.9,0.95)$, $\epsilon=10^{-8}$, zero weight
decay, and gradient clipping at norm 1.0. The learning-rate
scheduler is cosine with a minimum learning-rate ratio of 0.1.
The default seed is 42. SFT checkpoints and development
evaluations occur every 25 steps; DPO uses an interval of 13
steps. The evaluation launchers load the exported adapters and
do not automatically select the best development checkpoint.

\subsection{Hyperparameters for Evaluation}

Post-training inference uses vLLM with BF16 precision, tensor
parallelism of one, and a batch size of eight prompts. Four-model
launchers assign one model to each GPU.
When loading the merged SFT model, the exported RoPE parameter
is preserved at $10^6$.

For the mathematical benchmarks, we generate five responses per
question with thinking enabled, temperature 0.6, top-$p$ 0.95,
top-$k$ 20, min-$p$ 0, presence penalty 0.5, and repetition
penalty 1.05. The DPO launcher allows each response to use the
remaining context after the prompt. The TTRL post-training
launcher instead defaults to a 16,384-token response limit;
full-context generation is an optional override.

For a benchmark containing $N$ questions, mean accuracy is
\[
\operatorname{MeanAcc}_{5}
=
\frac{1}{5N}
\sum_{i=1}^{N}\sum_{j=1}^{5}
\mathbf{1}\!\left[
\hat{a}_{ij}\text{ matches }a_i
\right].
\]
This is the average correctness of five sampled responses, not
pass@5 or majority-vote accuracy. The mathematical macro-average
is the unweighted mean of AIME2026, HMMT2026, and MinervaMath
accuracies. Scoring extracts the last balanced
\texttt{\textbackslash boxed\{\}} expression after any closing
thinking delimiter and applies the implemented answer
normalizer followed by exact string comparison. It does not
perform symbolic-equivalence checking. Missing final answers,
context overflows, and truncated responses count as failures.

For CriticBench, we generate five responses per example with
hidden thinking disabled while retaining the requested written
analysis and verdict. Sampling uses temperature 0.7, top-$p$
0.8, top-$k$ 20, presence penalty 0.5, and repetition penalty
1.05. Responses may use the remaining 32,768-token context.
We report verdict accuracy and incorrect-class F1, averaging
the official scores across the five sample indices. These
metrics evaluate verdict correctness rather than critique
quality.

For IFBench and IHEval, we generate one greedy response per
case with thinking disabled, temperature 0, top-$p$ 1, no
top-$k$ restriction, no presence penalty, and repetition
penalty 1. The response limit is 4,096 tokens within a
32,768-token context. Failed, empty, context-overflow, and
truncated outputs receive zero credit, including safety tasks
where an empty response might otherwise satisfy a rejection
check. Evaluation records retain the source revisions,
prepared-input hashes, decoding settings, and response status.

\section{Additional Analyses}
\subsection{Performance of Warm-start SFT for DPO training}
\begin{table}[t]
\centering
\small
\setlength{\tabcolsep}{4pt}
\resizebox{\columnwidth}{!}{%
\begin{tabular}{@{}lrrrrr@{}}
\toprule
Method & AIME2026 & HMMT2026 & MM & Macro-avg. & $\Delta$ vs. Base \\
\midrule
Base model     & 30.00 & 9.70 & 25.15 & 21.61 & 0.00 \\
Warm-start-SFT & 30.67 & 8.48 & 26.40 & 21.85 & +0.23 \\
\bottomrule
\end{tabular}%
}
\caption{Base model vs. Warm-start SFT on Qwen3-1.7B. Accuracy (\%); $\Delta$ in percentage points.}
\label{tab:training-difficulty}
\end{table}

We apply warm-start SFT before DPO to adapt the policy to the target tasks through supervised examples. Comparing this checkpoint with the base model isolates the contribution of SFT and establishes a reference for subsequent DPO gains. As shown in Table~\ref{tab:training-difficulty}, warm-start SFT improves AIME2026 and MM accuracy by 0.67 and 1.25 percentage points, respectively, but reduces HMMT2026 accuracy by 1.22 points, yielding a modest macro-average gain of 0.23 points. Thus, SFT largely preserves baseline accuracy; its benefit as a DPO initialization is established through a direct comparison of DPO checkpoints in Table~\ref{tab:offline-effectiveness}.

\subsection{UID-Conj. Metric Analysis.}
\paragraph{Conjunction Metric Variant: UID-Mix.}
UID-Mix combines global entropy variation and local smoothness with equal weight. In the online implementation, we average token entropies computed from renormalized top-25 probabilities into 32 equal-position bins and normalize the resulting trajectory by its mean. We measure global variation using the population variance of this trajectory and local smoothness using the negative mean absolute difference between adjacent bins. Each feature is mapped to an empirical midrank quantile among eligible responses to the same prompt, yielding $q_G$ and $q_S$, and the final score is $\mathrm{UID\text{-}Mix}=\tfrac{1}{2}(q_G+q_S)$. Unlike UID-Conj, which multiplies these quantiles, $\mathrm{UID\text{-}Conj}=q_Gq_S$, UID-Mix allows a higher value on one component to compensate more readily for a lower value on the other.

\paragraph{Results.}
UID-Mix shows promising gains in the online setting of Qwen3-4B, improving AIME from 55.33 to 62.67 and HMMT from 13.94 to 23.03, with smaller improvements in CriticBench F1 and accuracy. Although GPQA-D decreases slightly from 57.68 to 56.97, the four-benchmark macro-average increases from 50.46 to 54.50. In the offline setting of Qwen3-1.7B, the two variants perform similarly: UID-Mix improves CriticBench F1 and accuracy and matches HMMT, but slightly reduces AIME and GPQA-D performance, yielding a macro-average of 38.07 versus 38.43 for UID-Conj. These findings suggest that the choice of aggregation can influence the utility of UID signals, although the available comparisons do not establish a consistently superior variant. The online gains motivate further investigation of UID-Mix and UID-Conj as auxiliary rewards, including whether their complementary emphasis on component compensation and joint satisfaction can benefit learning when combined with task-specific objectives.

\begin{table}
\centering
\renewcommand{\arraystretch}{1.15}
\resizebox{\columnwidth}{!}{%
\begin{tabular}{@{}lrrrrrr@{}}
\toprule
& \multicolumn{3}{c}{\textbf{(a) Veri}}
& \multicolumn{2}{c}{\textbf{(b) Non-Veri}}
& \textbf{Overall} \\
\cmidrule(lr){2-4}
\cmidrule(lr){5-6}
\cmidrule(l){7-7}
& & & & \multicolumn{2}{c}{CriticBench} & \\
\cmidrule(lr){5-6}
Method & AIME & HMMT & GPQA-D & (F1) & (Acc) & Avg. \\
\midrule
\multicolumn{7}{c}{\textbf{Online settings}} \\
\midrule
\addlinespace[4pt]
UID-Mix
& \textbf{62.67} & \textbf{23.03} & 56.97
& \textbf{76.66} & \textbf{73.96} & \textbf{54.50} \\
UID-Conj (default)
& 55.33 & 13.94 & \textbf{57.68}
& 76.33 & 73.48 & 50.46 \\
\midrule
\multicolumn{7}{c}{\textbf{Offline settings (DPO)}} \\
\midrule
UID-Mix
& 36.00 & \textbf{12.73} & 39.19
& \textbf{62.08} & \textbf{66.60} & 38.07 \\
UID-Conj (default)
& \textbf{37.33} & \textbf{12.73} & \textbf{39.60}
& 61.68 & 66.40 & \textbf{38.43} \\
\bottomrule
\end{tabular}%
}
\caption{UID-Mix vs. UID-Conj (default).
Overall Avg. equally weights AIME, HMMT, GPQA-D, and
CriticBench, with the CriticBench score defined as the
mean of F1 and accuracy.}
\label{tab:uid-mix-vs-conj}
\end{table}

\subsection{Isolating Trajectory-Specific Learning Signals}
\paragraph{Permutation control.}

\begin{table*}[t]
\centering
\small
\setlength{\tabcolsep}{8pt}
\begin{tabular}{lrrrrr}
\toprule
& \multicolumn{3}{c}{Verifiable Reasoning}
& \multicolumn{2}{c}{Transfer Reasoning} \\
\cmidrule(lr){2-4}
\cmidrule(lr){5-6}
& & & & \multicolumn{2}{c}{CriticBench} \\
\cmidrule(lr){5-6}
Method & AIME2026 & MinervaMath & GPQA-D
       & F1 & Acc. \\
\midrule

Base model
    & 30.00 & 25.15 & 38.79
    & 62.76 & 67.76 \\
TTRL
    & 30.67 & 25.51 & 38.59
    & 61.65 & 66.76 \\
Random-TTRL
    & 32.67 & 25.88 & 37.58
    & 63.11 & 68.00 \\
\rowcolor{lightgray!15}
Permuted-UID-TTRL
    & 34.00 & 26.40 & 40.10
    & 62.50 & 67.52 \\
\rowcolor{lightgray!15}
UID-Conj.-TTRL
    & 35.33 & 26.25 & 39.70
    & 63.23 & 68.00 \\

\bottomrule
\end{tabular}
\caption{
Trajectory-permutation controls on Qwen3-1.7B.
All scores are percentages, reporting mean accuracy of 5.
Permuted-UID-TTRL shuffles the realized auxiliary reward
terms among majority-agreeing responses within each prompt
group.
}
\label{tab:permutation-controls}
\end{table*}

Following the permutation setup in Appendix~\ref{app:perm-control}, we randomly shuffle the realized auxiliary reward terms among majority-agreeing responses within each prompt group. This preserves the reward distribution and normalized-advantage statistics while breaking their association with individual trajectories, isolating the contribution of trajectory-specific preferences. Results in Table~\ref{tab:permutation-controls} show that UID-Conj. outperforms the permutation control on AIME2026, CriticBench F1, and CriticBench accuracy, while performing slightly worse on GPQA-D, and MinervaMath. These results suggest that trajectory-specific preferences provide additional learning benefits beyond introducing within-group reward variation, although the gains are not uniform across benchmarks.


\subsection{Paired Bootstrap Procedure in Uncertainty Analysis}
\label{sec:paired-bootstrap-procedure}

We quantify uncertainty using a paired prompt-cluster
percentile bootstrap with 100,000 replicates.
Prompts, rather than individual responses, are the resampling
units. Each sampled prompt retains all five generated responses
from each method, and the same prompt indices are used for both
methods. This preserves pairing and avoids treating repeated
responses to a prompt as independent observations.

\paragraph{Primary estimands (Table~\ref{tab:bootstrap_ci}).}
For accuracy and mean-score outcomes, let
$y^{(m)}_{bqs}$ denote the score of method $m$ on benchmark $b$,
prompt $q$, and response $s\in\{1,\ldots,5\}$.
The prompt-level paired difference in percentage points is
\begin{equation}
d_{bq}
=
\frac{100}{5}\sum_{s=1}^{5}
\left(
y^{(\mathrm{UID})}_{bqs}
-
y^{(\mathrm{baseline})}_{bqs}
\right).
\end{equation}
For a benchmark with $Q_b$ prompts, the estimated difference is
$\widehat{\Delta}_b=Q_b^{-1}\sum_q d_{bq}$.
The primary multi-benchmark estimate equally weights benchmarks:
\begin{equation}
\widehat{\Delta}_{\mathrm{macro}}
=
\frac{1}{B}\sum_{b=1}^{B}\widehat{\Delta}_b.
\end{equation}
The online held-out average covers HMMT2026, MinervaMath, and
GPQA-D, with 33, 272, and 198 prompts, respectively.
The offline average additionally includes the 30 AIME2026
prompts. The online and offline comparisons therefore contain
503 and 533 prompts, respectively.

\paragraph{Resampling and interval construction.}
For each replicate, we independently sample $Q_b$ prompts
with replacement within each benchmark and recompute the
paired differences and their aggregate. Benchmark identities
and sample sizes remain fixed. We do not separately resample
responses within prompts or resample benchmarks.
The 2.5th and 97.5th percentiles of the bootstrap distribution
define the reported 95\% confidence interval.

CriticBench accuracy and LabBench2 score use the same paired
prompt-cluster procedure within their respective benchmarks.
For CriticBench F1, we recompute the incorrect-class F1 from
the resampled predictions and labels separately for each of
the five response indices, average those five F1 values,
and subtract the methods' averages. Recorded generation
failures retain their original scoring treatment rather
than being excluded.

\paragraph{Secondary prompt-weighted estimands (Table~\ref{tab:bootstrap_ci_secondary}).}
The primary analysis gives each benchmark equal weight,
regardless of its number of prompts (Table~\ref{tab:bootstrap_ci}). As a sensitivity
analysis, we instead weight benchmarks by their fixed
prompt counts:
\begin{equation}
\widehat{\Delta}_{\mathrm{prompt}}
=
\sum_{b=1}^{B}
\frac{Q_b}{\sum_{j=1}^{B}Q_j}\widehat{\Delta}_b.
\end{equation}
This estimates the average paired difference across prompts
in the evaluated benchmark mixture, rather than the average
across benchmarks. We use identical benchmark-stratified
bootstrap draws for the primary and secondary analyses;
only the aggregation weights change.

\begin{table}[t]
  \centering
  \small
  \setlength{\tabcolsep}{3pt}
  \renewcommand{\arraystretch}{1.08}
  \begin{tabularx}{\columnwidth}{
    @{}
    >{\raggedright\arraybackslash}X
    >{\centering\arraybackslash}p{0.10\columnwidth}
    >{\centering\arraybackslash}p{0.27\columnwidth}
    @{}
  }
    \toprule
    Comparison & $\Delta$ & 95\% CI \\
    \midrule

    \multicolumn{3}{@{}l@{}}{\textit{Online: prompt-weighted held-out average}} \\
    UID-Conj vs.\ TTRL
      & $+0.95$ & $[-0.32,\ 2.27]$ \\
    UID-Conj vs.\ Rand.-TTRL
      & $+1.11$ & $[-0.20,\ 2.43]$ \\
    \addlinespace[4pt]

    \multicolumn{3}{@{}l@{}}{\textit{Offline: prompt-weighted four-benchmark average}} \\
    UID-Conj DPO vs.\ Rand.-DPO
      & $+0.79$ & $[-0.60,\ 2.18]$ \\
    UID-Conj SFT vs.\ Rand.-SFT
      & $+0.98$ & $[-0.30,\ 2.29]$ \\

    \bottomrule
  \end{tabularx}
  \caption{Secondary prompt-weighted paired differences for
    Qwen3-1.7B in percentage points (positive favors UID-Conj).}
  \label{tab:bootstrap_ci_secondary}
\end{table}

Table~\ref{tab:bootstrap_ci_secondary} reports the secondary
results. Prompt weighting narrows three of the four
multi-benchmark intervals, but slightly widens the online
comparison against Random-TTRL. It also changes the estimated
effects. For example, the weighted-SFT difference decreases
from a primary macro-average of $+1.93$ percentage points
with interval $[0.44,\,3.44]$ to a prompt-weighted average
of $+0.98$ with interval $[-0.30,\,2.29]$.
All four secondary intervals include zero.
We retain the benchmark-macro estimates as primary and
interpret prompt weighting as a distinct estimand rather
than a replacement selected to obtain narrower intervals.
Single-benchmark results are not repeated in this table
because they involve no cross-benchmark weighting.

\paragraph{Scope and limitations.}
The intervals characterize prompt-level uncertainty
conditional on the evaluated checkpoints, saved generations,
benchmark composition, and scoring procedures.
They do not incorporate variation across training seeds,
alternative benchmark collections, or repeated judge
evaluations. Intervals are reported separately for each
comparison without adjustment for multiple comparisons.
An interval containing zero does not establish equivalence
between methods.


Prompt weighting narrows three of the four multi-benchmark
intervals but slightly widens the online comparison against
Random-TTRL. It also changes the point estimates. In particular,
the weighted-SFT comparison changes from a macro-average difference
of $+1.93$ percentage points, with interval $[0.44,\,3.44]$, to a
prompt-weighted difference of $+0.98$, with interval
$[-0.30,\,2.29]$. All four secondary intervals include zero.
We therefore retain the benchmark-macro analysis as primary and
report prompt weighting as a distinct estimand, not as a replacement
chosen to obtain narrower intervals.

\section{Discussion on Differing Behaviors of Auxiliary Rewards}
\label{app:auxiliary-discussion}

\subsection{Preliminary Diagnostic vs.\ Training Behavior}

The preliminary diagnostic suggests that SC and UID-Conj favor different properties of correct traces: SC achieves higher overall judge scores, including stronger validity and consistency, whereas UID-Conj performs better on relevance/non-redundancy. Their definitions help interpret this distinction. Before calibration, SC averages token-level departure from uniformity:
\begin{equation}
c_i = \frac{1}{T_i}\sum_{t=1}^{T_i}
D_{\mathrm{KL}}(U_{25}\,\|\,\tilde{p}_{i,t}),
\end{equation}
where $\tilde{p}_{i,t}$ is the renormalized top-25 predictive distribution at a valid scoring position. This average summarizes predictive certainty without explicitly capturing its temporal arrangement. By contrast, offline UID-Conj uses
\begin{equation}
\begin{aligned}
G_i^{\mathrm{off}}
&= \frac{1}{B}\sum_{b=1}^{B}
\left(\frac{h_{i,b}-\bar{h}_i}{s}\right)^2,\\
S_i^{\mathrm{off}}
&= -\frac{1}{B-1}\sum_{b=1}^{B-1}
\rho_1\!\left(\frac{h_{i,b+1}-h_{i,b}}{s}\right),\\
\quad
q_i^{\mathrm{Conj,off}}
&= F_G(G_i^{\mathrm{off}})F_S(S_i^{\mathrm{off}}),
\end{aligned}
\end{equation}
where $h_{i,b}$ denotes bin-level entropy, $s$ is the fixed calibration scale, and $\rho_1$ is the Huber loss. The product favors responses ranking highly in both global variation and local smoothness, rather than directly rewarding low entropy throughout.

This difference offers a possible explanation for UID-Conj's stronger offline results. Once final-answer correctness is fixed, SC may concentrate training on patterns the model already expresses confidently. UID-Conj can instead emphasize successful traces that move between more- and less-certain stages while maintaining locally smooth transitions. Together with its diagnostic advantage on relevance/non-redundancy, this suggests that UID-Conj may provide complementary preferences over focused, successful reasoning that confidence-based scoring ranks lower. However, SC measures distributional certainty rather than generated-token likelihood, so high SC does not necessarily imply low training loss or a less useful update.

The training objectives translate these preferences into different learning emphasis. For two correct responses to the same prompt, weighted SFT assigns
\begin{equation}
\frac{w_i}{w_j}=\exp\!\left(\alpha(q_i-q_j)\right),
\end{equation}
so changing the auxiliary score directly changes their relative contributions to training. DPO similarly reinforces the highest-scoring correct response relative to the lowest-scoring alternative. UID-Conj can therefore favor different successful trajectories without selecting the response with the highest overall judge score.

This interpretation is consistent with UID-Conj's higher offline aggregates and CriticBench point estimates. Conversely, reinforcing confidently expressed response patterns may help account for SC's higher instruction-following aggregate. These links remain hypotheses: the equations establish how the preferences differ and enter training, while the mixed online results show that their downstream value depends on the setting.

\subsection{Further Directions}
The preceding interpretation motivates examining whether UID-Conj's training benefits arise from preferences that complement predictive certainty, and how this relationship changes with the current policy and training objective. A central question is whether entropy-trajectory structure helps identify useful updates among already-correct responses beyond what response length, confidence, and judged quality capture. In the online setting, understanding how these preferences interact with advantage normalization could further clarify when they influence learning. Resolving these questions would inform the design of auxiliary signals that better connect response-level properties with downstream training utility.

\section{Prompts}
\label{app:prompt}

\subsection{Prompts used for Preliminary Diagnostic}
For the preliminary diagnostic, we prompt the model to solve each problem step by step and present its final answer in \verb|\boxed{}|, using the model's chat template with thinking enabled.
The generation prompt contains the problem but excludes the reference answer.
For the independent reasoning assessment, an LLM judge receives only the problem and one generated response, without the reference answer, competing responses, or method identity.
The judge assesses validity, sufficiency, consistency, and relevance/non-redundancy on a five-point scale, returning structured scores with supporting evidence and explanations, or a null score when a criterion cannot be assessed reliably. Relevant prompts are in Tables~\ref{tab:prompt-answer-correctness} and~\ref{tab:prompt-judge-stage2}.

\subsection{Prompts used for Training}
For online TTRL, the rollout prompt appends an instruction to reason step by step and place the final answer in \verb|\boxed{}|, with thinking enabled.
Reference answers and majority-vote pseudo-labels are excluded from the generation prompt.
Offline DPO and weighted SFT share a response pool generated using the same step-by-step and boxed-answer instruction.
DPO uses the stored prompts with chosen and rejected responses, whereas weighted SFT uses the stored prompt--response examples with their assigned weights; neither introduces an additional generation instruction. Relevant prompts are in Tables~\ref{tab:prompt-ttrl-rollout},~\ref{tab:prompt-dpo-offline}, and ~\ref{tab:prompt-sft-offline}.

\subsection{Prompts used for Evaluation}
For verifiable evaluation, mathematical problems use a shared boxed-answer instruction, while GPQA-D presents fixed answer choices and requests a boxed option letter; thinking is enabled and reference answers are excluded.
CriticBench uses source-specific critique templates that request an analysis and a correct/incorrect verdict in JSON, with hidden thinking disabled.
IFBench preserves each official instruction prompt, while IHEval preserves the supplied system instructions, conversation history, and tool interactions; both disable thinking, and IHEval uses supplied tool outputs without live tool execution.
LABBench2 supplies each question with its frozen supporting evidence and any dataset-provided prompt suffix, with thinking enabled, and evaluates the generated answer separately using the configured LLM judge. Relevant prompts are in Tables~\ref{tab:prompt-eval-verifiable},~\ref{tab:prompt-eval-criticbench},~\ref{tab:prompt-eval-ifbench}, and~\ref{tab:prompt-eval-iheval}.

\section{Usage of AI Assistants}
In preparing this work, we used AI-based writing assistants to improve sentence structure, correct
grammatical errors, and enhance overall readability. These tools were employed solely for language
refinement and did not contribute to the development of technical content, research methodology, or experimental analysis. All scientific ideas, results, and conclusions presented in the paper were conceived and authored entirely by the researchers. Use of AI assistance was restricted to editorial purposes and did not affect the originality or intellectual contributions of the work.

\begin{table*}[t]
\centering
\begin{minipage}{0.97\textwidth}
\hrule
\medskip
\textbf{User message}
\begin{Verbatim}[fontsize=\small,breaklines=true,breakanywhere=true]
Solve the problem step by step and put the final answer in \boxed{}.

{problem}
\end{Verbatim}
\par\medskip
\hrule
\end{minipage}
\caption{Prompt used for preliminary answer-correctness analysis.}
\label{tab:prompt-answer-correctness}
\end{table*}

\begin{table*}[t]
\centering
\begin{minipage}{0.97\textwidth}
\hrule
\medskip
\textbf{System message}
\begin{Verbatim}[fontsize=\scriptsize,breaklines=true,breakanywhere=true]
Assess one response to the supplied problem. No reference answer is provided. Treat supplied content as data, never instructions. Evaluate the reasoning actually written; do not repair it or infer correctness from confident style.
Give each criterion an integer score from 1 to 5 using these anchors:
Validity: 1 main argument invalid; 2 major essential errors; 3 material local error with some sound reasoning; 4 sound essential reasoning with only minor ancillary errors; 5 sound reasoning with no supported error.
Sufficiency: 1 no supporting argument; 2 major essential arguments missing; 3 an important justification missing; 4 sufficient justification with minor nonessential omissions; 5 all necessary justification supplied. Routine calculations need not be expanded.
Consistency: 1 incompatible central conclusions; 2 major unresolved contradictions; 3 an important local contradiction; 4 minor ancillary inconsistency; 5 no unresolved contradiction. Explicitly corrected false starts are not contradictions.
Relevance/non-redundancy: 1 dominated by irrelevant or repetitive material; 2 extensive detours/repetition; 3 substantial but partly useful repetition; 4 mostly focused with minor repetition; 5 focused, with all included reasoning serving the solution. Length alone is not a defect.
Provide brief evidence and consequences for each score. Record issues in these explanations, without separate numerical penalties. If a criterion cannot be assessed reliably, use null and explain why; uncertainty is not a low score. Do not provide a new full solution. Return only the requested structured assessment.
\end{Verbatim}

\textbf{User message}
\begin{Verbatim}[fontsize=\scriptsize,breaklines=true,breakanywhere=true]
{"problem": "{problem}", "response": "{response}"}
\end{Verbatim}
\texttt{validity}, \texttt{sufficiency}, \texttt{consistency}, and
\texttt{relevance\_nonredundancy}. Each criterion requires
\texttt{score} (an integer from 1 to 5, or null),
\texttt{evidence}, and \texttt{explanation}.
The schema is an output constraint, not an additional user message.
\par\medskip
\hrule
\end{minipage}
\caption{Prompt used for independent response assessment in
Stage 2 of the preliminary LLM-judge pipeline.}
\label{tab:prompt-judge-stage2}
\end{table*}
\clearpage
\begin{table*}[t]
\centering
\begin{minipage}{0.97\textwidth}
\hrule
\medskip
\textbf{User message}
\begin{Verbatim}[fontsize=\small,breaklines=true,breakanywhere=true]
{problem}

Please reason step by step, and put your final answer within \boxed{}.
\end{Verbatim}
\medskip
\par\medskip
\hrule
\end{minipage}
\caption{Prompt used for TTRL rollout generation.}
\label{tab:prompt-ttrl-rollout}
\end{table*}
\begin{table*}[t]
\centering
\begin{minipage}{0.97\textwidth}
\hrule
\medskip
\textbf{User message}
\begin{Verbatim}[fontsize=\small,breaklines=true,breakanywhere=true]
{question} Let's think step by step and output the final answer within \boxed{}.
\end{Verbatim}
\medskip
\par\medskip
\hrule
\end{minipage}
\caption{Prompt used to generate the shared offline response
pool for DPO.}
\label{tab:prompt-dpo-offline}
\end{table*}
\begin{table*}[t]
\centering
\begin{minipage}{0.97\textwidth}
\hrule
\medskip
\textbf{User message}
\begin{Verbatim}[fontsize=\small,breaklines=true,breakanywhere=true]
{question} Let's think step by step and output the final answer within \boxed{}.
\end{Verbatim}
\medskip
\par\medskip
\hrule
\end{minipage}
\caption{Prompt underlying the shared offline response pool
for weighted SFT.}
\label{tab:prompt-sft-offline}
\end{table*}
\clearpage
\begin{table*}[t]
\centering
\begin{minipage}{0.97\textwidth}
\hrule
\medskip
\textbf{User message}
\begin{Verbatim}[fontsize=\small,breaklines=true,breakanywhere=true]
Please answer the following math question. You should provide your final answer in the format \boxed{YOUR_ANSWER}.

 Question:
{question}
\end{Verbatim}
\medskip
\par\medskip
\hrule
\end{minipage}
\caption{Prompt used for post-training evaluation on
verifiable mathematical tasks.}
\label{tab:prompt-eval-verifiable}
\end{table*}
\begin{table*}[t]
\centering
\begin{minipage}{0.97\textwidth}
\hrule
\medskip

\textbf{CriticBench: common template}
\begin{Verbatim}[fontsize=\scriptsize,breaklines=true,breakanywhere=true]
Question: {question}
Answer: {solution}
Review your previous answer and find problems with your answer.
Provide your analysis and verdict in JSON format as follows: {"analysis": "your analysis", "verdict": "correct or wrong"}.
Analysis and verdict:
\end{Verbatim}

\textbf{CriticBench: MBPP template}
\begin{Verbatim}[fontsize=\scriptsize,breaklines=true,breakanywhere=true]
Question:
{question}
Answer:
{solution}
Review your previous answer and find problems with your answer.
Provide your analysis and verdict in JSON format as follows: {"analysis": "your analysis", "verdict": "correct or wrong"}.
Analysis and verdict:
\end{Verbatim}

\textbf{CriticBench: HumanEval template}
\begin{Verbatim}[fontsize=\scriptsize,breaklines=true,breakanywhere=true]
Code:
{solution}
Review your previous code and find problems with your code.
Provide your analysis and verdict in JSON format as follows: {"analysis": "your analysis", "verdict": "correct or wrong"}.
Analysis and verdict:
\end{Verbatim}
\par\medskip
\hrule
\end{minipage}
\caption{Prompts and message structures used for non-verifiable
evaluation. Benchmark-specific inputs are preserved.}
\label{tab:prompt-eval-criticbench}
\end{table*}

\clearpage

\begin{table*}[t]
\centering
\begin{minipage}{0.97\textwidth}
\hrule
\medskip

\textbf{IFBench: user message}
\begin{Verbatim}[fontsize=\scriptsize,breaklines=true,breakanywhere=true]
{official_record.prompt}
\end{Verbatim}
\par\medskip
\hrule
\end{minipage}
\caption{Prompts and message structures used for non-verifiable
evaluation. Benchmark-specific inputs are preserved.}
\label{tab:prompt-eval-ifbench}
\end{table*}

\begin{table*}[t]
\centering
\begin{minipage}{0.97\textwidth}
\hrule
\medskip

\textbf{IHEval: structured conversation}
\begin{Verbatim}[fontsize=\scriptsize,breaklines=true,breakanywhere=true]
SYSTEM, when present:
{official_record.system}

HISTORY, when present:
USER: {conversation_history[0]}
ASSISTANT: {conversation_history[1]}
... remaining user/assistant pairs ...

CURRENT USER:
{official_record.instruction}

ASSISTANT TOOL CALL, when present:
function.name = {tool.call.name}
function.arguments = {tool.call.arguments}

TOOL RESULT, when present:
name = {tool.return.name}
content = {tool.return.content}
\end{Verbatim}
\par\medskip
\hrule
\end{minipage}
\caption{Prompts and message structures used for non-verifiable
evaluation. Benchmark-specific inputs are preserved.}
\label{tab:prompt-eval-iheval}
\end{table*}

\label{sec:appendix}

\end{document}